%% file: main.tex
\pdfoutput=1

\documentclass[11pt]{article}

\usepackage{naacl2021}

\usepackage{times}
\usepackage{latexsym}

\usepackage[T1]{fontenc}

\usepackage[utf8]{inputenc}

\usepackage{microtype}

\input{symbols}

\input{libs}

\title{Double Perturbation: On the Robustness of Robustness and Counterfactual Bias Evaluation}

\author{
  Chong Zhang \quad Jieyu Zhao \quad Huan Zhang \quad Kai-Wei Chang \quad Cho-Jui Hsieh \\
  Department of Computer Science, UCLA \\
  {{\{chongz, jyzhao, kwchang, chohsieh\}@cs.ucla.edu},\quad huan@huan-zhang.com}
}

\begin{document}

\maketitle
\begin{abstract}
Robustness and counterfactual bias are usually evaluated on a test dataset. However, are these evaluations robust?
If the test dataset is perturbed slightly, will the evaluation results keep the same?
In this paper, we propose a ``double perturbation'' framework to uncover model weaknesses beyond the test dataset. The framework first perturbs the test dataset to construct abundant natural sentences similar to the test data, and then diagnoses the prediction change regarding a single-word substitution.
We apply this framework to study two perturbation-based approaches that are used to analyze models' robustness and counterfactual bias in English.
(1) For robustness, we focus on synonym substitutions and identify vulnerable examples where prediction can be altered.
Our proposed attack attains high success rates ($96.0\%$\textendash$99.8\%$) in finding vulnerable examples on both original and robustly trained CNNs and Transformers.
(2) For counterfactual bias, we focus on substituting demographic tokens (e.g., gender, race) and measure the shift of the \emph{expected} prediction among constructed sentences.
Our method is able to reveal the hidden model biases not directly shown in the test dataset.
Our code is available at \url{https://github.com/chong-z/nlp-second-order-attack}.
\end{abstract}

\section{Introduction}
Recent studies show that NLP models are vulnerable to adversarial perturbations.
A seemingly \qt{invariance transformation} (a.k.a. adversarial perturbation) such as synonym substitutions~\citep{alzantot-etal-2018-generating, Zang_2020} or syntax-guided paraphrasing~\citep{Iyyer2018AdversarialEG,huang2021generating} can alter the prediction.
To mitigate the model vulnerability, robust training methods have been proposed and shown effective~\citep{Miyato2017AdversarialTM, Jia2019CertifiedRT, Huang2019Achieving,zhou2020defense}.

\begin{figure}[!t]
 \centering
 \input{figures/intro}
 \caption{A vulnerable example beyond the test dataset. Numbers on the bottom right are the sentiment predictions for \w{film} and \w{movie}. \cblue $x_0$ comes from the test dataset and its prediction cannot be altered by the substitution $\w{film} \rightarrow \w{movie}$ (robust).
 \cyellow example $\tilde{x}_0$ is slightly perturbed but remains natural. Its prediction can be altered by the substitution (vulnerable).}
 \label{fig:intro}
\end{figure}
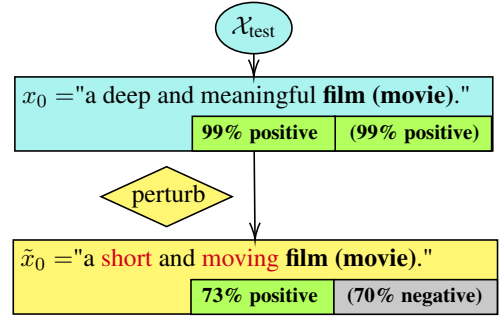

In most studies, model robustness is evaluated based on a given test dataset or synthetic sentences constructed from templates~\citep{Marco2020checklist}.
Specifically, the robustness of a model is often evaluated by the ratio of test examples where the model prediction cannot be altered by semantic-invariant perturbation. 
We refer to this type of evaluations as the \emph{\fo} robustness evaluation.
However, even if a model is \fo robust on an input sentence $x_0$, it is possible that the model is not robust on a natural sentence $\tilde{x}_0$ that is slightly modified from $x_0$. 
In that case, adversarial examples still exist even if \fo attacks cannot find any of them from the given test dataset.
Throughout this paper, we call $\tilde{x}_0$ a \emph{vulnerable example}. The existence of such examples exposes weaknesses in models' understanding and presents challenges for model deployment. 
\cref{fig:intro} illustrates an example.

\begin{figure}[!t]
 \centering
 \input{figures/circle}
 \caption{An illustration of the decision boundary. Diamond area denotes invariance transformations.
 \cblue $x_0$ is a robust input example (the entire diamond is green). \cyellow $\tilde{x}_0$ is a \emph{vulnerable} example in the neighborhood of $x_0$.
 \cred $\tilde{x}'_0$ is an \emph{adversarial} example to $\tilde{x}_0$. Note: $\tilde{x}'_0$ is \emph{not} an adversarial example to $x_0$ since they have different meanings to human (outside the diamond).}
 \label{fig:circle}
\end{figure}
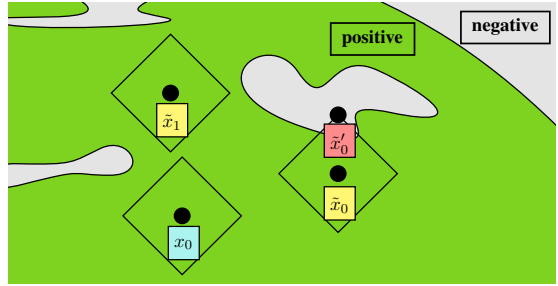

In this paper, we propose the \emph{double perturbation framework} for evaluating a stronger notion of \emph{\so robustness}.
Given a test dataset, we consider a model to be \emph{\so} robust if there is no vulnerable example that can be identified in the \emph{neighborhood} of given test instances (\cref{sec:neighbor}).
In particular, our framework first perturbs the test set to construct the neighborhood, and then diagnoses the robustness regarding a single-word synonym substitution.
Taking \cref{fig:circle} as an example, the model is \fo robust on the input sentence $x_0$ (the prediction cannot be altered), but it is not \so robust due to the existence of the vulnerable example $\tilde{x}_0$. Our framework is designed to identify $\tilde{x}_0$.

We apply the proposed framework and quantify \so robustness through two \emph{\so attacks} (\cref{sec:robustness}).
We experiment with English sentiment classification on the SST-2 dataset~\citep{socher-etal-2013-recursive} across various model architectures.
Surprisingly, although robustly trained CNN~\citep{Jia2019CertifiedRT} and Transformer~\citep{xu2020automatic} can achieve high robustness under strong attacks~\citep{alzantot-etal-2018-generating,Garg2020BAEBA} ($23.0\%$\textendash$71.6\%$ success rates), for around $96.0\%$ of the test examples our attacks can find a vulnerable example by perturbing 1.3 words on average. This finding indicates that these robustly trained models, despite being \fo robust, are not second-order robust.

Furthermore, we extend the double perturbation framework to evaluate counterfactual biases~\citep{Kusner2017Counterfactual} (\cref{sec:bias}) in English.
When the test dataset is small, our framework can help improve the evaluation robustness by revealing the hidden biases not directly shown in the test dataset.
Intuitively, a fair model should make the same prediction for nearly identical examples referencing different groups~\citep{Garg2019Counterfactual} with different protected attributes (e.g., gender, race).
In our evaluation, we consider a model \emph{biased} if substituting tokens associated with protected attributes changes the \emph{expected} prediction, which is the average prediction among all examples within the neighborhood.
For instance, a toxicity classifier is biased if it tends to increase the toxicity if we substitute $\w{straight} \rightarrow \w{gay}$ in an input sentence~\citep{dixon2018measuring}.
In the experiments, we evaluate the expected sentiment predictions on pairs of protected tokens (e.g., (\w{he}, \w{she}), (\w{gay}, \w{straight})), and demonstrate that our method is able to reveal the hidden model biases.

Our main contributions are: (1) We propose the double perturbation framework to diagnose the robustness of existing robustness and fairness evaluation methods. (2) We propose two \so attacks to quantify the stronger notion of \so robustness and reveal the models' vulnerabilities that cannot be identified by previous attacks. (3) We propose a counterfactual bias evaluation method to reveal the hidden model bias based on our double perturbation framework.

\section{The Double Perturbation Framework}
\label{sec:double-perturbation}

In this section, we describe the double perturbation framework which focuses on identifying vulnerable examples within a small neighborhood of the test dataset. The framework consists of a neighborhood perturbation and a word substitution. We start with defining word substitutions.

\subsection{Existing Word Substitution Strategy}
\label{sec:single-word}

We focus our study on word-level substitution, where existing works evaluate robustness and counterfactual bias by directly perturbing the test dataset. For instance, adversarial attacks alter the prediction by making synonym substitutions, and the fairness literature evaluates counterfactual fairness by substituting protected tokens.
We integrate the word substitution strategy into our framework as the component for evaluating robustness and fairness.

For simplicity, we consider a single-word substitution and denote it with the operator $\oplus$.
Let $\cX \subseteq \cV^l$ be the input space where $\cV$ is the vocabulary and $l$ is the sentence length, $\p = (p^{(1)}, p^{(2)}) \in \cV^2$ be a pair of synonyms (called \emph{patch words}), $\cX_{\p} \subseteq \cX$ denotes sentences with a single occurrence of $p^{(1)}$ (for simplicity we skip other sentences),
$x_0 \in \cX_{\p}$ be an input sentence, then $x_0 \oplus \p$ means ``substitute $p^{(1)} \rightarrow p^{(2)}$ in $x_0$''. The result after substitution is:
\begin{equation*}
x_0' = x_0 \oplus \p.
\end{equation*}
Taking \cref{fig:intro} as an example, where $\p =($\w{film}, \w{movie}$)$ and $x_0 =$ \s{a deep and meaningful film}, the perturbed sentence is $x_0' =$ \s{a deep and meaningful movie}.
Now we introduce other components in our framework.

\subsection{Proposed Neighborhood Perturbation}
\label{sec:neighbor}

Instead of applying the aforementioned word substitutions directly to the original test dataset, our framework perturbs the test dataset within a small neighborhood to construct similar natural sentences.
This is to identify vulnerable examples with respect to the model.
Note that examples in the neighborhood are not required to have the same meaning as the original example, since we only study the prediction difference caused by applying synonym substitution $\p$ (\cref{sec:single-word}).

\define{Constraints on the neighborhood.} We limit the neighborhood sentences within a small $\ell_0$ norm ball (regarding the test instance) to ensure syntactic similarity, and empirically ensure the naturalness through a language model. The neighborhood of an input sentence $x_0 \in \cX$ is:
\begin{equation}
\label{eq:aug}
\aug_{k}(x_0) \subseteq \text{Ball}_{k}(x_0) \cap \cX_\text{natural}, 
\end{equation}
where $\text{Ball}_{k}(x_0) = \{ x \mid \norm{x - x_0}_0 \leq k, x \in \cX \}$ is the $\ell_0$ norm ball around $x_0$ (i.e., at most $k$ different tokens), and $\cX_\text{natural}$ denotes natural sentences that satisfy a certain language model score which will be discussed next.

\define{Construction with masked language model.} We construct neighborhood sentences from $x_0$ by substituting at most $k$ tokens. As shown in \cref{alg:aug}, the construction employs a recursive approach and replaces one token at a time.
For each recursion, the algorithm first masks each token of the input sentence (may be the original $x_0$ or the $\tilde{x}$ from last recursion) separately and predicts likely replacements with a masked language model (e.g., DistilBERT, \citealt{Sanh2019DistilBERTAD}). To ensure the naturalness, we keep the top $20$ tokens for each mask with the largest logit (subject to a threshold, \cref{lst:line:logit}). Then, the algorithm constructs neighborhood sentences by replacing the mask with found tokens.
We use the notation $\tilde{x}$ in the following sections to denote the constructed sentences within the neighborhood.

\begin{algorithm}[h]
\SetAlgoLined
\KwData{Input sentence $x_0$, masked language model $\text{LM}$, max distance $k$.}
\SetKwFunction{FAug}{$\aug_{k}$}
\SetKwProg{Fn}{Function}{:}{end}
\Fn{\FAug{$x_0$}}{
    \lIf{$k = 0$}{
      \Return{$\{x_0\}$}
    }
    \If{$k \geq 2$}{
      \Return{$\bigcup_{\tilde{x} \in \aug_1(x_0)} \aug_{k - 1}(\tilde{x})$}\;
    }
    $\cX_\text{neighbor} \leftarrow \emptyset$\;
    \For{$i \leftarrow 0, \dots, \text{len}(x_0)-1$}{
        $T, L \leftarrow \text{LM.fillmask}(x_0, i)$\;
        \Comment{Mask $i_\text{th}$ token and return candidate tokens and corresponding logits.}
        $L \leftarrow \text{SortDecreasing}(L)$\;
        $l_\text{min} \leftarrow \max\{L^{(\kappa)},\; L^{(0)} - \delta\}$\; \label{lst:line:logit}
        \Comment{$L^{(i)}$ denotes the $i_\text{th}$ element. We empirically set $\kappa \leftarrow 20$ and $\delta \leftarrow 3$.}
        $T_\text{new} \leftarrow \{ t \mid l > l_\text{min},\; (t, l) \in T \times L \}$\;
        $\cX_\text{new} \leftarrow \{ x_0 \mid x_0^{(i)} \leftarrow t,\; t \in T_\text{new}\}$\;
        \Comment{Construct new sentences by replacing the $i_\text{th}$ token.}
        $\cX_\text{neighbor} \leftarrow \cX_\text{neighbor} \cup \cX_\text{new}$\;
    }
    \Return{$\cX_\text{neighbor}$}\;
}
\caption{Neighborhood construction }
\label{alg:aug}
\end{algorithm}

\section{Evaluating \SO Robustness}
\label{sec:robustness}

\begin{figure*}[!t]
    \centering
    \input{figures/attack-flow}
    \caption{The attack flow for \sob (\cref{alg:soft}). \cblue $x_0$ is the input sentence and \cyellow $\tilde{x}_0$ is our constructed vulnerable example (the prediction can be altered by substituting $\w{film}\rightarrow \w{movie}$).
    \cgreen boxes in the middle show intermediate sentences, and $f_\text{soft}(x)$ denotes the probability outputs for \w{film} and \w{movie}.}
    \label{fig:attack-flow}
\end{figure*}
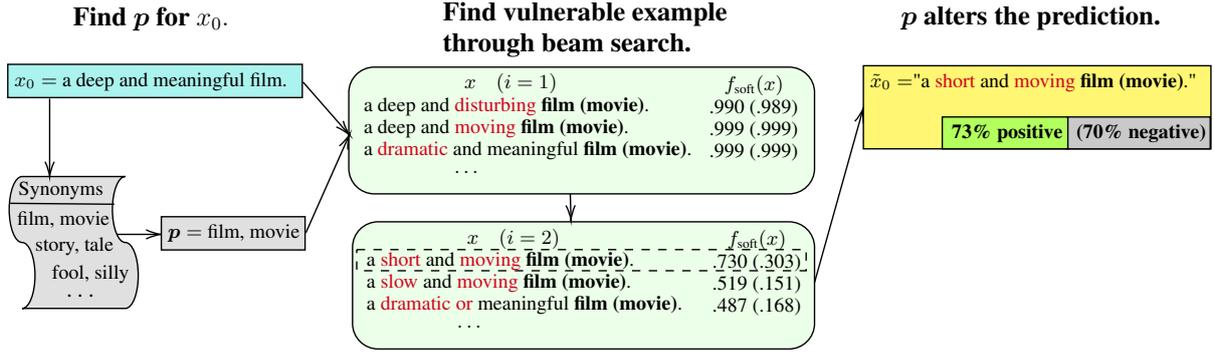

With the proposed double perturbation framework, we design two black-box attacks\footnote{Black-box attacks only observe the model outputs and do not know the model parameters or the gradient.} to identify vulnerable examples within the neighborhood of the test set.
We aim at evaluating the robustness for inputs beyond the test set.

\subsection{Previous \FO Attacks}
\label{sec:fo-attacks}

Adversarial attacks search for small and invariant perturbations on the model input that can alter the prediction. 
To simplify the discussion, in the following, we take a binary classifier $f(x) : \cX \rightarrow \{0, 1\}$ as an example to describe our framework.
Let $x_0$ be the sentence from the test set with label $y_0$, then the smallest perturbation $\delta^*$ under $\ell_0$ norm distance is:\footnote{For simplicity, we use $\ell_0$ norm distance to measure the similarity, but other distance metrics can be applied.} 
\begin{equation*}
\begin{split}
\delta^* := \argmin_\delta \norm{\delta}_0 \st f(x_0 \oplus \delta) \ne y_0.
\end{split}
\end{equation*}
Here $\delta = \p_1 \oplus \dots \oplus \p_l$ denotes a series of substitutions. In contrast, our \so attacks fix $\delta = \p$ and search for the vulnerable $x_0$.

\subsection{Proposed \SO Attacks}
\label{sec:so-attacks}
\So attacks study the prediction difference caused by applying $\p$. For notation convenience we define the prediction difference $F(x; \p) : \cX \times \cV^2 \rightarrow \{-1, 0, 1\}$ by:\footnote{We assume a binary classification task, but our framework is general and can be extended to multi-class classification.}
\begin{equation}
\label{eq:F}
F(x; \p) := f(x \oplus \p) - f(x).
\end{equation}
Taking \cref{fig:intro} as an example, the prediction difference for 
$\tilde{x}_0$ on $\p$ is $F(\tilde{x}_0; \p)  = f($\s{\dots moving \textbf{movie}.}$) - f($\s{\dots moving \textbf{film}.}$) = -1$.

Given an input sentence $x_0$, we want to find patch words $\p$ and a vulnerable example $\tilde{x}_0$ such that $f(\tilde{x}_0 \oplus \p) \ne f(\tilde{x}_0)$.
Follow \citet{alzantot-etal-2018-generating}, we choose $\p$ from a predefined list of counter-fitted synonyms~\citep{mrksic-etal-2016-counter} that maximizes $|f_\text{soft}(p^{(2)}) - f_\text{soft}(p^{(1)})|$. Here $f_\text{soft}(x) : \cX \rightarrow [0, 1]$ denotes probability output (e.g., after the softmax layer but before the final argmax), $f_\text{soft}(p^{(1)})$ and $f_\text{soft}(p^{(2)})$ denote the predictions for the single word, and we enumerate through all possible $\p$ for $x_0$.
Let $k$ be the neighborhood distance, then the attack is equivalent to solving:
\begin{equation}
\label{eq:hard}
\tilde{x}_0 = \argmax_{x \in \aug_k(x_0)} |F(x; \p)|.
\end{equation}

\definep{Brute-force attack}{(\textit{\soe}).} A naive approach for solving \cref{eq:hard} is to enumerate through $\aug_k(x_0)$. The enumeration finds the smallest perturbation, but is only applicable for small $k$ (e.g., $k \leq 2$) given the exponential complexity.

\definep{Beam-search attack}{(\textit{\sob}).} The efficiency can be improved by utilizing the probability output, where we solve \cref{eq:hard} by minimizing the cross-entropy loss with regard to $x \in \aug_k(x_0)$:
\begin{align}
\label{eq:loss}
\cL(x; \p) := -\log(1 - f_\text{min}) - \log(f_\text{max}),
\end{align}
where $f_\text{min}$ and $f_\text{max}$ are the smaller and the larger output probability between $f_\text{soft}(x)$ and $f_\text{soft}(x \oplus \p)$, respectively.
Minimizing \cref{eq:loss} effectively leads to $f_\text{min} \rightarrow 0$ and $f_\text{max} \rightarrow 1$, and we use a beam search to find the best $x$. At each iteration, we construct sentences through $\aug_1(x)$ and only keep the top 20 sentences with the smallest $\cL(x; \p)$. 
We run at most $k$ iterations, and stop earlier if we find a vulnerable example.
We provide the detailed implementation in \cref{alg:soft} and a flowchart in \cref{fig:attack-flow}.

\begin{algorithm}[h]
\SetAlgoLined
\SetKwFunction{FSOAttack}{$\text{\SOAttack}_{k}$}
\KwData{Input sentence $x_0$, synonyms $\cP$, model functions $F$ and $f_\text{soft}$, loss $\cL$, max distance $k$.}
\SetKwFunction{FSOAttack}{$\text{\sob}_{k}$}
\SetKwProg{Fn}{Function}{:}{end}
\Fn{\FSOAttack{$x_0$}}{
    $\p \leftarrow \argmax\limits_{\p \in \cP \st x_0 \in \cX_{\p}} |f_\text{soft}(p^{(2)}) -  f_\text{soft}(p^{(1)})|$\;
    $\cX_\text{beam} \leftarrow \{x_0\}$\;
    \For{$i \leftarrow 1, \dots, k$}{
        $\cX_\text{new} \leftarrow \bigcup_{\tilde{x} \in \cX_\text{beam}} \aug_1(\tilde{x})$\;
        $\tilde{x}_0 \leftarrow \argmax_{x \in \cX_\text{new}} |F(x; \p)|$\;
        \lIf{$F(\tilde{x}_0; \p) \ne 0$}{\Return{$\tilde{x}_0$}}
        $\cX_\text{new} \leftarrow \text{SortIncreasing}(\cX_\text{new}, \cL)$\;
        $\cX_\text{beam} \leftarrow \{\cX_\text{new}^{(0)}, \dots, \cX_\text{new}^{(\beta-1)}\}$\;
        \Comment{Keep the best beam. We set $\beta \leftarrow 20$.}
    }
    \Return{None}\;
}
\caption{Beam-search attack (\sob)}
\label{alg:soft}
\end{algorithm}

\subsection{Experimental Results}
\label{sec:eob-exp}

In this section, we evaluate the \so robustness of existing models and show the quality of our constructed vulnerable examples.

\subsubsection{Setup}
\label{sec:exp-rob-setup}

\begin{table*}[!t]
    \centering
    \resizebox{\textwidth}{!}{\input{tables/attack_examples}}
    \caption{Sampled attack results on the robust BoW. For Genetic and BAE the goal is to find an \emph{adversarial} example that alters the original prediction, whereas for SO-Enum and SO-Beam the goal is to find a \emph{vulnerable} example beyond the test set such that the prediction can be altered by substituting $\w{bad} \rightarrow \w{unhealthy}$.}
    \label{tb:attack_examples}
\end{table*}

We follow the setup from the robust training literature \citep{Jia2019CertifiedRT, xu2020automatic} and experiment with both the base (non-robust) and robustly trained models.
We train the binary sentiment classifiers on the SST-2 dataset with bag-of-words (BoW), CNN, LSTM, and attention-based models.

\define{Base models.} For BoW, CNN, and LSTM, all models use pre-trained GloVe embeddings~\citep{pennington2014glove}, and have one hidden layer of the corresponding type with 100 hidden size. Similar to the baseline performance reported in GLUE~\citep{wang2019glue}, our trained models have an evaluation accuracy of 81.4\%, 82.5\%, and 81.7\%, respectively.
For attention-based models, we train a 3-layer Transformer (the largest size in \citealt{Shi2020Robustness}) and fine-tune a pre-trained \w{bert-base-uncased} from HuggingFace~\citep{wolf2020huggingfaces}.
The Transformer uses 4 attention heads and 64 hidden size, and obtains 82.1\% accuracy. The BERT-base uses the default configuration and obtains 92.7\% accuracy.

\define{Robust models (\fo).} With the same setup as base models, we apply robust training methods to improve the resistance to word substitution attacks.
\citet{Jia2019CertifiedRT} provide a provably robust training method through Interval Bound Propagation (IBP, \citealt{dvijotham2018IBP}) for all word substitutions on BoW, CNN and LSTM.
\citet{xu2020automatic} provide a provably robust training method on general computational graphs through a combination of forward and backward linear bound propagation, and the resulting 3-layer Transformer is robust to up to 6 word substitutions.
For both works we use the same set of counter-fitted synonyms provided in \citet{Jia2019CertifiedRT}.
We skip BERT-base due to the lack of an effective robust training method.

\define{Attack success rate (\fo).} We quantify \fo robustness through attack success rate, which measures the ratio of test examples that an \emph{adversarial} example can be found.
We use \fo attacks as a reference due to the lack of a direct baseline.
We experiment with two black-box attacks: (1) The Genetic attack~\citep{alzantot-etal-2018-generating, Jia2019CertifiedRT} uses a population-based optimization algorithm that generates both syntactically and semantically similar adversarial examples, by replacing words within the list of counter-fitted synonyms. (2) The BAE attack~\citep{Garg2020BAEBA} generates coherent adversarial examples by masking and replacing words using BERT.
For both methods we use the implementation provided by TextAttack~\citep{morris2020textattack}.

\define{Attack success rate (\so).} We also quantify \so robustness through attack success rate, which measures the ratio of test examples that a \emph{vulnerable} example can be found.
To evaluate the impact of neighborhood size, we experiment with two configurations: (1) For the small neighborhood ($k=2$), we use \soe that finds the most similar vulnerable example. (2) For the large neighborhood ($k=6$), \soe is not applicable and we use \sob to find vulnerable examples.
We consider the most challenging setup and use patch words $\p$ from the same set of counter-fitted synonyms as robust models (they are provably robust to these synonyms on the test set).
We also provide a random baseline to validate the effectiveness of minimizing \cref{eq:loss} (\cref{apd:random-baseline}).

\define{Quality metrics (perplexity and similarity).}
We quantify the quality of our constructed vulnerable examples through two metrics:
(1) GPT-2~\citep{radford2019language} perplexity quantifies the naturalness of a sentence (smaller is better). We report the perplexity for both the original input examples and the constructed vulnerable examples.
(2) $\ell_0$ norm distance quantifies the disparity between two sentences (smaller is better). We report the distance between the input and the vulnerable example.
Note that \fo attacks have different objectives and thus cannot be compared directly.

\subsubsection{Results}
\label{sec:exp-robustness}

We experiment with the validation split (872 examples) on a single RTX 3090. The average running time per example (in seconds) on base LSTM is 31.9 for Genetic, 1.1 for BAE, 7.0 for SO-Enum ($k=2$), and 1.9 for SO-Beam ($k=6$). We provide additional running time results in \cref{apd:runtime}.
\cref{tb:attack_examples} provides an example of the attack result where all attacks are successful (additional examples in \cref{apd:additional-robustness}).
As shown, our \so attacks find a vulnerable example by replacing \w{grease} $\rightarrow$ \w{musicals}, and the vulnerable example has different predictions for \w{bad} and \w{unhealthy}. Note that, Genetic and BAE have different objectives from \so attacks and focus on finding the adversarial example.
Next we discuss the results from two perspectives.

\begin{table}[!t]
    \centering
    \resizebox{.9\columnwidth}{!}{\input{tables/attack_success_rate}}
    \caption{The average rates over 872 examples (100 for Genetic due to long running time). \So attacks achieve higher successful rate since they are able to search beyond the test set. 
    }
    \label{tb:attack_success}
\end{table}

\define{\So robustness.} We observe that existing robustly trained models are not \so robust.
As shown in \cref{tb:attack_success}, our \so attacks attain high success rates not only on the base models but also on the robustly trained models. For instance, on the robustly trained CNN and Transformer, \sob finds vulnerable examples within a small neighborhood for around $96.0\%$ of the test examples,
even though these models have improved resistance to strong \fo attacks (success rates drop from $62.0\%$\textendash$74.3\%$ to $23.0\%$\textendash$71.6\%$ for Genetic and BAE).\footnote{BAE is more effective on robust models as it may use replacement words outside the counter-fitted synonyms.}
This phenomenon can be explained by the fact that both \fo attacks and robust training methods focus on synonym substitutions on the test set, whereas our attacks, due to their \so nature, find vulnerable examples beyond the test set, and the search is not required to maintain semantic similarity.
Our methods provide a way to further investigate the robustness (or find vulnerable and adversarial examples) even when the model is robust to the test set.

\define{Quality of constructed vulnerable examples.} As shown in \cref{tb:quality}, \so attacks are able to construct vulnerable examples by perturbing 1.3 words on average, with a slightly increased perplexity.
For instance, on the robustly trained CNN and Transformer, \sob constructs vulnerable examples by perturbing 1.3 words on average, with the median\footnote{We report median due to the unreasonably large perplexity on certain sentences. e.g., 395 for \s{that's a cheat.} but 6740 for \s{that proves perfect cheat.}} perplexity increased from around 165 to around 210.
We provide metrics for \fo attacks in \cref{apd:additional-robustness} as they have different objectives and are not directly comparable.

Furthermore, applying existing attacks on the vulnerable examples constructed by our method will lead to much smaller perturbations.
As a reference, on the robustly trained CNN, Genetic attack constructs adversarial examples by perturbing 2.7 words on average (starting from the input examples). However, if Genetic starts from our vulnerable examples, it would only need to perturb a single word (i.e., the patch words $\p$) to alter the prediction.
These results demonstrate the weakness of the models (even robustly trained) for those inputs beyond the test set.

\begin{table}[!t]
    \centering
    \resizebox{\columnwidth}{!}{\input{tables/quality_gpt2_l0}}
    \caption{The quality metrics for \so methods. We report the median perplexity (PPL) and average $\ell_0$ norm distance. The original PPL may differ across models since we only count successful attacks.}
    \label{tb:quality}
\end{table}

\subsubsection{Human Evaluation}

We perform human evaluation on the examples constructed by \sob.
Specifically, we randomly select 100 successful attacks and evaluate both the original examples and the vulnerable examples.
To evaluate the naturalness of the constructed examples, we ask the annotators to score the likelihood (on a Likert scale of 1-5, 5 to be the most likely) of being an original example based on the grammar correctness. To evaluate the semantic similarity after applying the synonym substitution $\p$, we ask the annotators to predict the sentiment of each example, and calculate the ratio of examples that maintain the same sentiment prediction after the synonym substitution. For both metrics, we take the median from 3 independent annotations.
We use US-based annotators on Amazon’s Mechanical Turk\footnote{\url{https://www.mturk.com}} and pay \$0.03 per annotation, and expect each annotation to take 10 seconds on average (effectively, the hourly rate is about \$11). See \cref{apd:human} for more details.

As shown in \cref{tb:human}, the naturalness score only drop slightly after the perturbation, indicating that our constructed vulnerable examples have similar naturalness as the original examples. As for the semantic similarity, we observe that 85\% of the original examples maintain the same meaning after the synonym substitution, and the corresponding ratio is 71\% for vulnerable examples. This indicates that the synonym substitution is an invariance transformation for most examples.

\begin{table}[!t]
    \centering
    \resizebox{.8\columnwidth}{!}{\input{tables/human_evaluation}}
    \caption{The quality metrics from human evaluation.}
    \label{tb:human}
\end{table}

\section{Evaluating Counterfactual Bias}
\label{sec:bias}

In addition to evaluating \so robustness, we further extend the double perturbation framework (\cref{sec:double-perturbation}) to evaluate counterfactual biases by setting $\p$ to pairs of protected tokens. We show that our method can reveal the hidden model bias.

\subsection{Counterfactual Bias}
\label{sec:sub-bias}

In contrast to \so robustness, where we consider the model vulnerable as long as there exists \emph{one} vulnerable example, counterfactual bias focuses on the \emph{expected} prediction, which is the average prediction among all examples within the neighborhood.
We consider a model biased if the expected predictions for protected groups are different (assuming the model is not intended to discriminate between these groups).
For instance, a sentiment classifier is biased if the expected prediction for inputs containing \w{woman} is more positive (or negative) than inputs containing \w{man}. 
Such bias is harmful as they may make unfair decisions based on protected attributes, for example in situations such as hiring and college admission.

\begin{figure}[!t]
 \centering
 \input{figures/bias}
 \caption{An illustration of an unbiased model vs. a biased model.
 \cgreen and gray indicate the probability of positive and negative predictions, respectively. \textbf{Left}: An unbiased model where the $(x, x \oplus \p)$ pair (\cyellow-\cred dots) is relatively parallel to the decision boundary.
 \textbf{Right}: A biased model where the predictions for $x \oplus \p$ (\cred) are usually more negative (gray) than $x$ (\cyellow).}
 \label{fig:bias-twoball}
\end{figure}
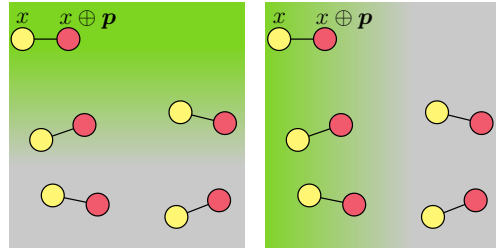

\define{Counterfactual token bias.} We study a narrow case of counterfactual bias, where counterfactual examples are constructed by substituting protected tokens in the input. A naive approach of measuring this bias is to construct counterfactual examples directly from the test set, however such evaluation may not be robust since test examples are only a small subset of natural sentences.
Formally, let $\p$ be a pair of protected tokens such as (\w{he}, \w{she}) or (\w{Asian}, \w{American}), $\cX_\text{test} \subseteq \cX_{\p}$ be a test set (as in \cref{sec:single-word}),
we define \emph{counterfactual token bias} by:
\begin{equation}
\label{eq:bg}
B_{\p, k} := \EEE_{x \in \aug_k(\cX_{\text{test}})} F_\text{soft}(x; \p).
\end{equation}
We calculate \cref{eq:bg} through an enumeration across all natural sentences within the neighborhood.\footnote{For gender bias, we employ a blacklist to avoid adding gendered tokens during the neighborhood construction. This is to avoid semantic shift when, for example, $\p=(\w{he}, \allowbreak \w{she})$ such that it may refer to different tokens after the substitution.}
Here $\aug_k(\cX_\text{test}) = \bigcup_{x \in \cX_\text{test}} \aug_k(x)$ denotes the union of neighborhood examples (of distance $k$) around the test set, and $F_\text{soft}(x; \p) : \cX \times \cV^2 \rightarrow [-1, 1]$ denotes the difference between probability outputs $f_\text{soft}$ (similar to \cref{eq:F}):
\begin{equation}
\label{eq:Fsoft}
F_\text{soft}(x; \p) := f_\text{soft}(x \oplus \p) - f_\text{soft}(x).
\end{equation} 
The model is unbiased on $\p$ if $B_{\p, k} \approx 0$, whereas a positive or negative $B_{\p, k}$ indicates that the model shows preference or against to $p^{(2)}$, respectively. \cref{fig:bias-twoball} illustrates the distribution of $(x, x \oplus \p)$ for both an unbiased model and a biased model.

The aforementioned neighborhood construction does not introduce additional bias.
For instance, let $x_0$ be a sentence containing \w{he}, even though it is possible for $\aug_1(x_0)$ to contain many stereotyping sentences (e.g., contains tokens such as \w{doctor} and \w{driving}) that affect the distribution of $f_\text{soft}(x)$, but it does not bias \cref{eq:Fsoft} as we only care about the prediction difference of replacing $\w{he}\rightarrow\w{she}$. The construction has no information about the model objective, thus it would be difficult to bias $f_\text{soft}(x)$ and $f_\text{soft}(x \oplus \p)$ differently.

\subsection{Experimental Results}
\label{sec:exp-bias}

In this section, we use gender bias as a running example, and demonstrate the effectiveness of our method by revealing the hidden model bias. We provide additional results in \cref{apd:additional-tokens}.

\begin{table}[!t]
 \centering
 \resizebox{.7\columnwidth}{!}{\input{tables/filtered_stats}}
 \caption{The number of original examples ($k=0$) and the number of perturbed examples ($k=3$) in $\cX_\text{filter}$.}
 \label{tb:filtered-stats}
\end{table}

\subsubsection{Setup}
We evaluate counterfactual token bias on the SST-2 dataset with both the base and debiased models.
We focus on binary gender bias and set $\p$ to pairs of gendered pronouns from \citet{zhao-etal-2018-gender}. 

\define{Base Model.} We train a single layer LSTM with pre-trained GloVe embeddings and 75 hidden size (from TextAttack, \citealt{morris2020textattack}).
The model has 82.9\% accuracy similar to the baseline performance reported in GLUE.

\define{Debiased Model.} Data-augmentation with gender swapping has been shown effective in mitigating gender bias~\citep{zhao-etal-2018-gender,zhao2019gender}. We augment the training split by swapping all male entities with the corresponding female entities and vice-versa. We use the same setup as the base LSTM and attain 82.45\% accuracy.

\define{Metrics.} We evaluate model bias through the proposed $B_{\p, k}$ for $k = 0, \dots, 3$. Here the bias for $k=0$ is effectively measured on the original test set, and the bias for $k\geq1$ is measured on our constructed neighborhood. We randomly sample a subset of constructed examples when $k=3$ due to the exponential complexity.

\define{Filtered test set.} To investigate whether our method is able to reveal model bias that was hidden in the test set, we construct a filtered test set on which the bias cannot be observed directly.
Let $\cX_\text{test}$ be the original validation split, we construct $\cX_\text{filter}$ by the equation below and empirically set $\epsilon = 0.005$. We provide statistics in \cref{tb:filtered-stats}.
\begin{equation*}
\cX_\text{filter} := \{ x \mid |F_\text{soft}(x; \p)| < \epsilon ,\; x \in \cX_\text{test} \}.
\end{equation*}

\subsubsection{Results}

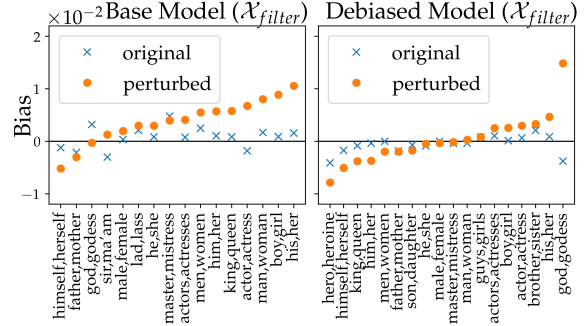
\begin{figure}[!t]
 \centering
 \input{figures/ctb-per-token}
 \caption{Our proposed $B_{\p, k}$ measured on $\cX_\text{filter}$. Here ``original'' is equivalent to $k=0$, ``perturbed'' is equivalent to $k=3$, $\p$ is in the form of $(\w{male}, \w{female})$.
 }
 \label{fig:bias-errorbar}
\end{figure}

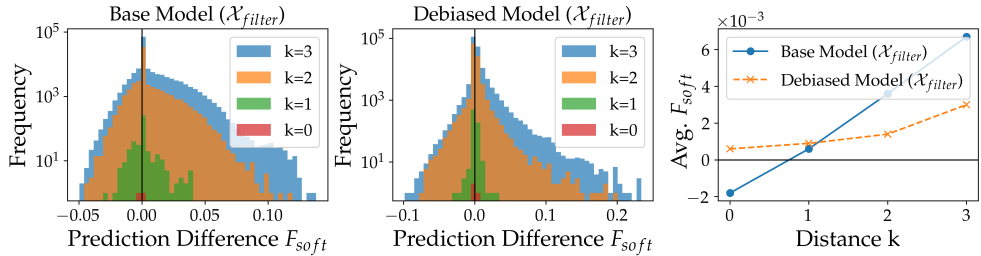
\begin{figure*}[!t]
 \centering
 \input{figures/hidden-bias}
 \caption{\textbf{Left and Middle:} Histograms for $F_\text{soft}(x; \p)$ (x-axis) with $\p = (\w{actor}, \w{actress})$.
 \textbf{Right:} The plot for the average $F_\text{soft}(x; \p)$ (i.e., counterfactual token bias) vs. neighborhood distance $k$. Results show that the counterfactual bias on $\p$ can be revealed when increasing $k$.
 \vspace{-10pt}}
 \label{fig:bias-histogram}
\end{figure*}

Our method is able to reveal the hidden model bias on $\cX_\text{filter}$, which is not visible with naive measurements.
In \cref{fig:bias-errorbar}, the naive approach ($k=0$) observes very small  biases on most tokens (as constructed). In contrast, when evaluated by our double perturbation framework ($k=3$), we are able to observe noticeable bias, where most $\p$ has a positive bias on the base model. This observed bias is in line with the measurements on the original $\cX_\text{test}$ (\cref{apd:additional-tokens}), indicating that we reveal the correct model bias. Furthermore, we observe mitigated biases in the debiased model, which demonstrates the effectiveness of data augmentation.

To demonstrate how our method reveals hidden bias, we conduct a case study with $\p = (\w{actor},  \allowbreak \w{actress})$ and show the relationship between the bias $B_{\p, k}$ and the neighborhood distance $k$.
We present the histograms for \fsp{} in \cref{fig:bias-histogram} and plot the corresponding $B_{\p, k}$ vs.\ $k$ in the right-most panel. Surprisingly, for the base model, the bias is negative when $k=0$, but becomes positive when $k=3$.
This is because the naive approach only has two test examples (\cref{tb:filtered-stats}) thus the measurement is not robust. In contrast, our method is able to construct \num{141780} similar natural sentences when $k=3$ and shifts the distribution to the right (positive). As shown in the right-most panel, the bias is small when $k=1$, and becomes more significant as $k$ increases (larger neighborhood).
As discussed in \cref{sec:sub-bias}, the neighborhood construction does not introduce additional bias, and these results demonstrate the effectiveness of our method in revealing hidden model bias.

\section{Related Work}
\define{\Fo robustness evaluation.} A line of work has been proposed to study the vulnerability of natural language models, through transformations such as
character-level perturbations~\citep{ebrahimi-etal-2018-hotflip},
word-level perturbations~\citep{jin2019bert,ren-etal-2019-generating, Yang2019greedy,hsieh2019robustness,Cheng_2020, li-etal-2020-bert-attack},
prepending or appending a sequence~\citep{Jia2017AdversarialEF, Wallace2019UniversalAT},
and generative models~\citep{zhao2018generating}.
They focus on constructing adversarial examples from the test set that alter the prediction, whereas our methods focus on finding vulnerable examples beyond the test set whose predictions can be altered.

\define{Robustness beyond the test set.} Several works have studied model robustness beyond test sets but mostly focused on computer vision tasks. \citet{zhang2018the} demonstrate that a robustly trained model could still be vulnerable to small perturbations if the input comes from a distribution only slightly different than a normal test set (e.g., images with slightly different contrasts). \citet{hendrycks2019benchmarking} study more sources of common corruptions such as brightness, motion blur and fog. Unlike in computer vision where simple image transformations can be used, in our natural language setting, generating a valid example beyond test set is more challenging because language semantics and grammar must be maintained.

\define{Counterfactual fairness.} \citet{Kusner2017Counterfactual} propose \emph{counterfactual fairness} and consider a model fair if changing the protected attributes does not affect the distribution of prediction.
We follow the definition and focus on evaluating the counterfactual bias between pairs of protected tokens.
Existing literature quantifies fairness on a test dataset or through templates~\citep{feldman2015certifying, kiritchenko-mohammad-2018-examining, may-etal-2019-measuring,huang2019reducing}. For instance,
\citet{Garg2019Counterfactual} quantify the absolute counterfactual token fairness gap on the test set;
\citet{prabhakaran-etal-2019-perturbation} study perturbation sensitivity for named entities on a given set of corpus. \citet{wallace-etal-2019-universal,sheng2019woman,sheng2020towards} study how language generation models respond differently to prompt sentences containing mentions of different demographic groups.
In contrast, our method quantifies the bias on the constructed neighborhood.

\section{Conclusion}
This work proposes the double perturbation framework to identify model weaknesses beyond the test dataset, and study a stronger notion of robustness and counterfactual bias. We hope that our work can stimulate the research on further improving the robustness and fairness of natural language models.

\section*{Acknowledgments}
We thank anonymous reviewers for their helpful feedback. We thank UCLA-NLP group for the valuable discussions and comments. The research is supported NSF \#1927554, \#1901527, \#2008173 and \#2048280 and an Amazon Research Award. 


\section*{Ethical Considerations}

\define{Intended use.} One primary goal of NLP models is the generalization to real-world inputs. However, existing test datasets and templates are often not comprehensive, and thus it is difficult to evaluate real-world performance~\citep{pmlr-v97-recht19a, Marco2020checklist}.
Our work sheds a light on quantifying performance for inputs beyond the test dataset and help uncover model weaknesses prior to the real-world deployment.

\define{Misuse potential.} Similar to other existing adversarial attack methods~\citep{ebrahimi-etal-2018-hotflip, jin2019bert, zhao2018generating}, our \so attacks can be used for finding vulnerable examples to a NLP system. Therefore, it is essential to study how to improve the robustness of NLP models against second-order attacks.

\define{Limitations.} While the core idea about the double perturbation framework is general, in \cref{sec:bias}, we consider only binary gender in the analysis of counterfactual fairness due to the restriction of the English corpus we used, which only have words associated with binary gender such as \w{he/she}, \w{waiter/waitress}, etc. 

\bibliography{anthology,references}
\bibliographystyle{acl_natbib}

\clearpage

\appendix

\input{appendix}

\end{document}

%% file: symbols.tex
\usepackage{graphicx,amssymb,amsmath,bm}

\newcommand{\bp}{\bm{p}}

\newcommand{\cL}{\mathcal{L}}

\newcommand{\cP}{\mathcal{P}}

\newcommand{\cV}{\mathcal{V}}

\newcommand{\cX}{\mathcal{X}}

\newcommand{\EE}{\mathbb{E}}

\newcommand{\argmin}{\mathop{\mathrm{argmin}}}
\newcommand{\argmax}{\mathop{\mathrm{argmax}}}

\let\emptyset\varnothing


%% file: libs.tex
\usepackage{amssymb}
\usepackage{bbm}
\usepackage{hyperref}
\usepackage{mathtools}
\usepackage{tikz, pgfplots}
\usepackage{listings}
\usepackage{xcolor}
\usepackage{booktabs}
\usepackage{multirow}
\usepackage{multicol}
\usepackage{makecell}
\usepackage{xspace}
\usepackage{tabularx}
\usepackage[htt]{hyphenat}
\usepackage[textsize=small]{todonotes}
\usepackage[group-separator={,}]{siunitx}

\usepackage{physics}
\usepackage{amsmath}
\usepackage{tikz}
\usepackage{mathdots}
\usepackage{yhmath}
\usepackage{cancel}
\usepackage{color}
\usepackage{array}
\usepackage{multirow}
\usepackage{amssymb}
\usepackage{gensymb}
\usepackage{tabularx}
\usepackage{booktabs}
\usetikzlibrary{fadings}
\usetikzlibrary{patterns}
\usetikzlibrary{shadows.blur}
\usetikzlibrary{shapes}

\usepackage[ruled,vlined,linesnumbered,noend]{algorithm2e}
\SetArgSty{textup}
\SetAlFnt{\small}
\SetKwRepeat{Do}{do}{while}

\SetCommentSty{mycommfont}
\SetKwComment{Comment}{$\triangleright$\ }{}

\definecolor{xyellow}{rgb}{0.97, 0.91, 0.11}
\definecolor{xblue}{rgb}{0.67, 0.95, 0.93}
\definecolor{xred}{rgb}{0.82,0.01,0.11}

\definecolor{xtyellow}{rgb}{0.83, 0.69, 0.22}
\definecolor{xtblue}{rgb}{0.0, 0.45, 0.73}
\definecolor{xtred}{rgb}{0.81, 0.09, 0.13}
\definecolor{xtgreen}{rgb}{0.0, 0.5, 0.0}

\definecolor{aogreen}{rgb}{0.0, 0.5, 0.0}

\newcommand\hmm[1]{\ifnum\ifhmode\spacefactor\else2000\fi>1000 \uppercase{#1}\else#1\fi}

\newcommand\cyellow{\textcolor{xtyellow}{\hmm{y}ellow}\xspace}
\newcommand\cblue{\textcolor{xtblue}{\hmm{b}lue}\xspace}
\newcommand\cred{\textcolor{xtred}{\hmm{r}ed}\xspace}
\newcommand\cgreen{\textcolor{xtgreen}{\hmm{g}reen}\xspace}

\usepackage[nameinlink,capitalize]{cleveref}
\crefformat{section}{\S#2#1#3} 
\crefformat{subsection}{\S#2#1#3}
\crefformat{subsubsection}{\S#2#1#3}

\newcommand{\st}{\text{ s.t. }}

\renewcommand{\norm}[1]{\left\lVert#1\right\rVert}

\newcommand{\EEE}{\mathop\EE}

\newcommand{\w}[1]{\texttt{#1}}
\newcommand{\s}[1]{\texttt{#1}}
\newcommand{\qt}[1]{\text{``#1''}}

\newcommand{\first}[1]{#1}

\newcommand{\define}[1]{\noindent\textbf{#1}\enskip}
\newcommand{\definep}[2]{\noindent\textbf{#1}\;#2\enskip}

\newcommand{\fo}{first-order\xspace}
\newcommand{\so}{second-order\xspace}

\newcommand{\Fo}{First-order\xspace}
\newcommand{\So}{Second-order\xspace}

\newcommand{\FO}{First-Order\xspace}
\newcommand{\SO}{Second-Order\xspace}

\newcommand{\soe}{SO-Enum\xspace}
\newcommand{\sob}{SO-Beam\xspace}

\newcommand{\fsp}{$F_\text{soft}(x; \p)$}
\newcommand{\p}{\bp}

\newcommand{\aug}{\text{Neighbor}}

\newcommand\adv[1]{\textcolor{xtred}{#1}}

\newcommand\posp[1]{\colorbox{xtgreen!30}{#1}}
\newcommand\negp[1]{\colorbox{gray!30}{#1}}



%% file: figures/intro.tex
\tikzset{every picture/.style={line width=0.75pt}} 

\resizebox{.85\columnwidth}{!}{
\begin{tikzpicture}[x=0.75pt,y=0.75pt,yscale=-1,xscale=1, every node/.style={scale=1.1}]

\draw  [fill={rgb, 255:red, 255; green, 247; blue, 115 }  ,fill opacity=1 ] (102.58,110.5) -- (143.83,129.5) -- (102.58,148.5) -- (61.33,129.5) -- cycle ;

\draw  [fill={rgb, 255:red, 171; green, 243; blue, 238 }  ,fill opacity=1 ]  (7,53.5) -- (308,53.5) -- (308,100.5) -- (7,100.5) -- cycle  ;
\draw (10,57.5) node [anchor=north west][inner sep=0.75pt]  [font=\normalsize] [align=left] {$\displaystyle x_{0} =$"a deep and meaningful \textbf{film (movie)}."\\};
\draw  [fill={rgb, 255:red, 171; green, 243; blue, 238 }  ,fill opacity=1 ]  (157.5, 22.5) circle [x radius= 24.04, y radius= 16.26]   ;
\draw (141.5,13.4) node [anchor=north west][inner sep=0.75pt]  [font=\normalsize]  {$\mathcal{X}_{\text{test}}$};
\draw  [fill={rgb, 255:red, 255; green, 247; blue, 115 }  ,fill opacity=1 ]  (6,157) -- (311,157) -- (311,204) -- (6,204) -- cycle  ;
\draw (9,161) node [anchor=north west][inner sep=0.75pt]  [font=\normalsize] [align=left] {$\displaystyle \tilde{x}_{0} =$"a \textcolor[rgb]{0.82,0.01,0.11}{short} and \textcolor[rgb]{0.82,0.01,0.11}{moving} \textbf{film (movie)}." \ \ \ \ \ \ \\};
\draw  [color={rgb, 255:red, 0; green, 0; blue, 0 }  ,draw opacity=1 ][fill={rgb, 255:red, 178; green, 255; blue, 93 }  ,fill opacity=1 ]  (117.93,182.17) -- (207.93,182.17) -- (207.93,204.17) -- (117.93,204.17) -- cycle  ;
\draw (154.93,193.17) node  [font=\small] [align=left] {\begin{minipage}[lt]{58.341892pt}\setlength\topsep{0pt}
\begin{flushright}
\textcolor[rgb]{0,0,0}{\textbf{73\% positive}}
\end{flushright}

\end{minipage}};
\draw  [color={rgb, 255:red, 0; green, 0; blue, 0 }  ,draw opacity=1 ][fill={rgb, 255:red, 201; green, 201; blue, 201 }  ,fill opacity=1 ]  (207.92,182.17) -- (310.92,182.17) -- (310.92,204.17) -- (207.92,204.17) -- cycle  ;
\draw (251.42,193.17) node  [font=\small] [align=left] {\begin{minipage}[lt]{67.011892pt}\setlength\topsep{0pt}
\begin{flushright}
\textcolor[rgb]{0,0,0}{\textbf{(70\% negative)}}
\end{flushright}

\end{minipage}};
\draw  [color={rgb, 255:red, 0; green, 0; blue, 0 }  ,draw opacity=1 ][fill={rgb, 255:red, 178; green, 255; blue, 93 }  ,fill opacity=1 ][line width=0.75]   (208.27,78.5) -- (307.27,78.5) -- (307.27,100.5) -- (208.27,100.5) -- cycle  ;
\draw (251.42,89.5) node  [font=\small] [align=left] {\begin{minipage}[lt]{64.451216pt}\setlength\topsep{0pt}
\begin{flushright}
\textcolor[rgb]{0,0,0}{\textbf{(99\% positive)}}
\end{flushright}

\end{minipage}};
\draw  [color={rgb, 255:red, 0; green, 0; blue, 0 }  ,draw opacity=1 ][fill={rgb, 255:red, 178; green, 255; blue, 93 }  ,fill opacity=1 ]  (118.77,78.5) -- (208.77,78.5) -- (208.77,100.5) -- (118.77,100.5) -- cycle  ;
\draw (154.93,89.5) node  [font=\small] [align=left] {\begin{minipage}[lt]{58.341892pt}\setlength\topsep{0pt}
\begin{flushright}
\textcolor[rgb]{0,0,0}{\textbf{99\% positive}}
\end{flushright}

\end{minipage}};
\draw (79,120.5) node [anchor=north west][inner sep=0.75pt]   [align=left] {{perturb}};
\draw    (157.5,38.76) -- (157.5,51.5) ;
\draw [shift={(157.5,53.5)}, rotate = 270] [color={rgb, 255:red, 0; green, 0; blue, 0 }  ][line width=0.75]    (10.93,-3.29) .. controls (6.95,-1.4) and (3.31,-0.3) .. (0,0) .. controls (3.31,0.3) and (6.95,1.4) .. (10.93,3.29)   ;
\draw    (157.73,100.5) -- (158.25,155) ;
\draw [shift={(158.27,157)}, rotate = 269.45] [color={rgb, 255:red, 0; green, 0; blue, 0 }  ][line width=0.75]    (10.93,-3.29) .. controls (6.95,-1.4) and (3.31,-0.3) .. (0,0) .. controls (3.31,0.3) and (6.95,1.4) .. (10.93,3.29)   ;

\end{tikzpicture}

}

%% file: figures/circle.tex
\tikzset{every picture/.style={line width=.75pt}}

\resizebox{.95\columnwidth}{!}{
\begin{tikzpicture}[x=0.75pt,y=0.75pt,yscale=-1,xscale=1]
\clip (0,20) rectangle (424,239.5);

\draw  [draw opacity=0][fill={rgb, 255:red, 126; green, 211; blue, 33 }  ,fill opacity=1 ] (0,0) -- (434,0) -- (434,269.5) -- (0,269.5) -- cycle ;
\draw  [fill={rgb, 255:red, 228; green, 228; blue, 228 }  ,fill opacity=1 ] (197.5,-53) .. controls (217.5,-63) and (698.5,12) .. (599.5,96) .. controls (500.5,180) and (537.5,241) .. (465.5,159) .. controls (393.5,77) and (337.5,43) .. (296.5,23) .. controls (255.5,3) and (177.5,-43) .. (197.5,-53) -- cycle ;
\draw  [fill={rgb, 255:red, 228; green, 228; blue, 228 }  ,fill opacity=1 ] (7.5,146) .. controls (51.5,147) and (48.5,145) .. (71.5,134) .. controls (94.5,123) and (110.5,155) .. (78.5,155) .. controls (46.5,155) and (62,149) .. (11.5,162) .. controls (-39,175) and (-36.5,145) .. (7.5,146) -- cycle ;
\draw  [fill={rgb, 255:red, 228; green, 228; blue, 228 }  ,fill opacity=1 ] (224.5,74) .. controls (248.5,110) and (247.5,55) .. (286.5,73) .. controls (325.5,91) and (363,120) .. (296.5,106) .. controls (230,92) and (317.5,148) .. (227.5,117) .. controls (137.5,86) and (200.5,38) .. (224.5,74) -- cycle ;
\draw  [fill={rgb, 255:red, 228; green, 228; blue, 228 }  ,fill opacity=1 ] (47.5,14) .. controls (-8.5,-13) and (176.5,-14) .. (132,7) .. controls (87.5,28) and (95.5,26) .. (122.5,32) .. controls (149.5,38) and (4.5,44) .. (-9.5,35) .. controls (-23.5,26) and (103.5,41) .. (47.5,14) -- cycle ;
\draw   (134.75,140.8) -- (180.7,186.75) -- (134.75,232.7) -- (88.8,186.75) -- cycle ;
\draw  [fill={rgb, 255:red, 0; green, 0; blue, 0 }  ,fill opacity=1 ] (128.68,186.75) .. controls (128.68,183.4) and (131.4,180.68) .. (134.75,180.68) .. controls (138.1,180.68) and (140.82,183.4) .. (140.82,186.75) .. controls (140.82,190.1) and (138.1,192.82) .. (134.75,192.82) .. controls (131.4,192.82) and (128.68,190.1) .. (128.68,186.75) -- cycle ;

\draw   (255.75,107.8) -- (301.7,153.75) -- (255.75,199.7) -- (209.8,153.75) -- cycle ;
\draw  [fill={rgb, 255:red, 0; green, 0; blue, 0 }  ,fill opacity=1 ] (249.68,107.8) .. controls (249.68,104.45) and (252.4,101.73) .. (255.75,101.73) .. controls (259.1,101.73) and (261.82,104.45) .. (261.82,107.8) .. controls (261.82,111.15) and (259.1,113.87) .. (255.75,113.87) .. controls (252.4,113.87) and (249.68,111.15) .. (249.68,107.8) -- cycle ;
\draw  [fill={rgb, 255:red, 0; green, 0; blue, 0 }  ,fill opacity=1 ] (249.68,153.75) .. controls (249.68,150.4) and (252.4,147.68) .. (255.75,147.68) .. controls (259.1,147.68) and (261.82,150.4) .. (261.82,153.75) .. controls (261.82,157.1) and (259.1,159.82) .. (255.75,159.82) .. controls (252.4,159.82) and (249.68,157.1) .. (249.68,153.75) -- cycle ;
\draw   (125.75,44.8) -- (171.7,90.75) -- (125.75,136.7) -- (79.8,90.75) -- cycle ;
\draw  [fill={rgb, 255:red, 0; green, 0; blue, 0 }  ,fill opacity=1 ] (119.68,90.75) .. controls (119.68,87.4) and (122.4,84.68) .. (125.75,84.68) .. controls (129.1,84.68) and (131.82,87.4) .. (131.82,90.75) .. controls (131.82,94.1) and (129.1,96.82) .. (125.75,96.82) .. controls (122.4,96.82) and (119.68,94.1) .. (119.68,90.75) -- cycle ;


\draw  [fill={rgb, 255:red, 171; green, 243; blue, 238 }  ,fill opacity=1 ]  (124,195) -- (148,195) -- (148,221) -- (124,221) -- cycle  ;
\draw (127,203) node [anchor=north west][inner sep=0.75pt]   [align=left] {$\displaystyle x_{0}$};

\draw  [fill={rgb, 255:red, 255; green, 247; blue, 115 }  ,fill opacity=1 ]  (245,164) -- (269,164) -- (269,190) -- (245,190) -- cycle  ;
\draw (248,171) node [anchor=north west][inner sep=0.75pt]   [align=left] {$\displaystyle \tilde{x}_{0}$};
\draw  [fill={rgb, 255:red, 255; green, 140; blue, 140 }  ,fill opacity=1 ]  (245,115) -- (269,115) -- (269,141) -- (245,141) -- cycle  ;
\draw (248,121) node [anchor=north west][inner sep=0.75pt]   [align=left] {$\displaystyle \tilde{x}'_{0}$};
\draw  [fill={rgb, 255:red, 255; green, 247; blue, 115 }  ,fill opacity=1 ]  (115,99) -- (139,99) -- (139,125) -- (115,125) -- cycle  ;
\draw (118,106) node [anchor=north west][inner sep=0.75pt]   [align=left] {$\displaystyle \tilde{x}_{1}$};
\draw  [line width=1]   (351,27) -- (420,27) -- (420,52) -- (351,52) -- cycle  ;
\draw (358,31) node [anchor=north west][inner sep=0.75pt]   [align=left] {\textbf{negative}};

\draw  [line width=1]   (250,36) -- (315,36) -- (315,61) -- (250,61) -- cycle  ;
\draw (257,40) node [anchor=north west][inner sep=0.75pt]   [align=left] {\textbf{positive}};

\end{tikzpicture}

}

%% file: figures/attack-flow.tex
\tikzset{every picture/.style={line width=0.75pt}} 

\resizebox{\textwidth}{!}{
\begin{tikzpicture}[x=0.75pt,y=0.75pt,yscale=-1,xscale=1, every node/.style={scale=1.2}]

\draw  [fill={rgb, 255:red, 224; green, 224; blue, 224 }  ,fill opacity=1 ] (97.21,145) .. controls (92.1,145) and (87.96,155.66) .. (87.96,168.8) .. controls (87.96,181.94) and (92.1,192.6) .. (97.21,192.6) .. controls (102.32,192.6) and (106.46,203.26) .. (106.46,216.4) .. controls (106.46,229.54) and (102.32,240.2) .. (97.21,240.2) -- (23.25,240.2) .. controls (28.35,240.2) and (32.49,229.54) .. (32.49,216.4) .. controls (32.49,203.26) and (28.35,192.6) .. (23.25,192.6) .. controls (18.14,192.6) and (14,181.94) .. (14,168.8) .. controls (14,155.66) and (18.14,145) .. (23.25,145) -- cycle ;
\draw    (14.67,164.33) -- (89.33,165) ;

\draw  [fill={rgb, 255:red, 236; green, 255; blue, 233 }  ,fill opacity=1 ] (257.33,84.3) .. controls (257.33,74.19) and (265.53,66) .. (275.63,66) -- (572.7,66) .. controls (582.81,66) and (591,74.19) .. (591,84.3) -- (591,139.2) .. controls (591,149.31) and (582.81,157.5) .. (572.7,157.5) -- (275.63,157.5) .. controls (265.53,157.5) and (257.33,149.31) .. (257.33,139.2) -- cycle ;
\draw  [fill={rgb, 255:red, 236; green, 255; blue, 233 }  ,fill opacity=1 ] (260,196.03) .. controls (260,185.89) and (268.22,177.67) .. (278.37,177.67) -- (572.63,177.67) .. controls (582.78,177.67) and (591,185.89) .. (591,196.03) -- (591,251.13) .. controls (591,261.28) and (582.78,269.5) .. (572.63,269.5) -- (278.37,269.5) .. controls (268.22,269.5) and (260,261.28) .. (260,251.13) -- cycle ;
\draw    (43,88.75) -- (43,143.25) ;
\draw [shift={(43,145.25)}, rotate = 270] [color={rgb, 255:red, 0; green, 0; blue, 0 }  ][line width=0.75]    (10.93,-3.29) .. controls (6.95,-1.4) and (3.31,-0.3) .. (0,0) .. controls (3.31,0.3) and (6.95,1.4) .. (10.93,3.29)   ;
\draw    (91,186.75) -- (121,186.75) ;
\draw [shift={(123,186.75)}, rotate = 180] [color={rgb, 255:red, 0; green, 0; blue, 0 }  ][line width=0.75]    (10.93,-3.29) .. controls (6.95,-1.4) and (3.31,-0.3) .. (0,0) .. controls (3.31,0.3) and (6.95,1.4) .. (10.93,3.29)   ;
\draw    (224.5,76.75) -- (256.69,113.74) ;
\draw [shift={(258,115.25)}, rotate = 228.97] [color={rgb, 255:red, 0; green, 0; blue, 0 }  ][line width=0.75]    (10.93,-3.29) .. controls (6.95,-1.4) and (3.31,-0.3) .. (0,0) .. controls (3.31,0.3) and (6.95,1.4) .. (10.93,3.29)   ;
\draw    (226.5,185.75) -- (257.19,116.08) ;
\draw [shift={(258,114.25)}, rotate = 473.78] [color={rgb, 255:red, 0; green, 0; blue, 0 }  ][line width=0.75]    (10.93,-3.29) .. controls (6.95,-1.4) and (3.31,-0.3) .. (0,0) .. controls (3.31,0.3) and (6.95,1.4) .. (10.93,3.29)   ;
\draw    (416,157.5) -- (416.45,174.75) ;
\draw [shift={(416.5,176.75)}, rotate = 268.51] [color={rgb, 255:red, 0; green, 0; blue, 0 }  ][line width=0.75]    (10.93,-3.29) .. controls (6.95,-1.4) and (3.31,-0.3) .. (0,0) .. controls (3.31,0.3) and (6.95,1.4) .. (10.93,3.29)   ;
\draw    (591,221.5) -- (625.46,97.43) ;
\draw [shift={(626,95.5)}, rotate = 465.52] [color={rgb, 255:red, 0; green, 0; blue, 0 }  ][line width=0.75]    (10.93,-3.29) .. controls (6.95,-1.4) and (3.31,-0.3) .. (0,0) .. controls (3.31,0.3) and (6.95,1.4) .. (10.93,3.29)   ;
\draw  [dash pattern={on 4.5pt off 4.5pt}] (263,197.75) -- (585,197.75) -- (585,213) -- (263,213) -- cycle ;

\draw  [fill={rgb, 255:red, 171; green, 243; blue, 238 }  ,fill opacity=1 ]  (13,64) -- (224,64) -- (224,88) -- (13,88) -- cycle  ;
\draw (16,68) node [anchor=north west][inner sep=0.75pt]  [font=\small] [align=left] {$\displaystyle x_{0} =$ a deep and meaningful film.};
\draw  [fill={rgb, 255:red, 224; green, 224; blue, 224 }  ,fill opacity=1 ]  (123,174) -- (226,174) -- (226,198) -- (123,198) -- cycle  ;
\draw (126,178) node [anchor=north west][inner sep=0.75pt]  [font=\small] [align=left] {$\displaystyle \p=$ film, movie};
\draw (268.33,180.67) node [anchor=north west][inner sep=0.75pt]  [font=\small] [align=left] { \ \ \ \ \ \ \ \ \ \ \ \ \ \ \ \ \ \ \ \ $\displaystyle x\ \ \ \ ( i=2)$\\a \textcolor[rgb]{0.82,0.01,0.11}{short} and \textcolor[rgb]{0.82,0.01,0.11}{moving} \textbf{film (movie)}.\\a \textcolor[rgb]{0.82,0.01,0.11}{slow} and \textcolor[rgb]{0.82,0.01,0.11}{moving} \textbf{film (movie)}.\\a \textcolor[rgb]{0.82,0.01,0.11}{dramatic} \textcolor[rgb]{0.82,0.01,0.11}{or} meaningful \textbf{film (movie)}.\\ \ \ \ \ \ \ \ \ \ \ \ \ \ \ \ \ \ \ $\displaystyle \cdots $};
\draw (516,182) node [anchor=north west][inner sep=0.75pt]  [font=\small] [align=left] {$\displaystyle \ \ \ f_{\text{soft}}( x)$\\.730 (.303)\\.519 (.151)\\.487 (.168)};
\draw (266.29,69) node [anchor=north west][inner sep=0.75pt]  [font=\small] [align=left] { \ \ \ \ \ \ \ \ \ \ \ \ \ \ \ \ \ \ \ \ $\displaystyle x\ \ \ \ ( i=1)$\\a deep and \textcolor[rgb]{0.82,0.01,0.11}{disturbing} \textbf{film (movie)}.\\a deep and \textcolor[rgb]{0.82,0.01,0.11}{moving} \textbf{film (movie)}.\\a \textcolor[rgb]{0.82,0.01,0.11}{dramatic} and meaningful \textbf{film (movie)}.\\ \ \ \ \ \ \ \ \ \ \ \ \ \ \ \ \ \ \ $\displaystyle \cdots $};
\draw (513.92,70.33) node [anchor=north west][inner sep=0.75pt]  [font=\small] [align=left] {$\displaystyle \ \ \ f_{\text{soft}}( x)$\\.990 (.989)\\.999 (.999)\\.999 (.999)};
\draw (18,168) node [anchor=north west][inner sep=0.75pt]  [font=\small] [align=left] {film, movie};
\draw (31,187) node [anchor=north west][inner sep=0.75pt]  [font=\small] [align=left] {story, tale};
\draw (43,207) node [anchor=north west][inner sep=0.75pt]  [font=\small] [align=left] {fool, silly};
\draw (53,226) node [anchor=north west][inner sep=0.75pt]   [align=left] {$\displaystyle \cdots $};
\draw (19,147) node [anchor=north west][inner sep=0.75pt]  [font=\small] [align=left] {$\displaystyle \text{Synonyms}$};
\draw (58,21) node [anchor=north west][inner sep=0.75pt]  [font=\large] [align=left] {\textbf{Find }$\displaystyle \p$\textbf{ for }$\displaystyle x_{0} .$};
\draw (323,17) node [anchor=north west][inner sep=0.75pt]  [font=\large] [align=left] {\textbf{Find vulnerable  example}\\\textbf{through beam search.}};
\draw (651,20) node [anchor=north west][inner sep=0.75pt]  [font=\large] [align=left] {$\displaystyle \p$\textbf{ alters the prediction.}};
\draw  [fill={rgb, 255:red, 255; green, 247; blue, 115 }  ,fill opacity=1 ]  (626,65) -- (875,65) -- (875,125) -- (626,125) -- cycle  ;
\draw (629,69) node [anchor=north west][inner sep=0.75pt]  [font=\small] [align=left] {$\displaystyle \tilde{x}_{0} =$"a \textcolor[rgb]{0.82,0.01,0.11}{short} and \textcolor[rgb]{0.82,0.01,0.11}{moving} \textbf{film (movie)}."\\\\};
\draw  [color={rgb, 255:red, 0; green, 0; blue, 0 }  ,draw opacity=1 ][fill={rgb, 255:red, 201; green, 201; blue, 201 }  ,fill opacity=1 ]  (771.5,102) -- (874.5,102) -- (874.5,125) -- (771.5,125) -- cycle  ;
\draw (817,114) node  [font=\small] [align=left] {\begin{minipage}[lt]{67.011892pt}\setlength\topsep{0pt}
\begin{flushright}
\textcolor[rgb]{0,0,0}{\textbf{(70\% negative)}}
\end{flushright}

\end{minipage}};
\draw  [color={rgb, 255:red, 0; green, 0; blue, 0 }  ,draw opacity=1 ][fill={rgb, 255:red, 178; green, 255; blue, 93 }  ,fill opacity=1 ]  (682.51,102) -- (772.51,102) -- (772.51,125) -- (682.51,125) -- cycle  ;
\draw (722,114) node  [font=\small] [align=left] {\begin{minipage}[lt]{58.341892pt}\setlength\topsep{0pt}
\begin{flushright}
\textcolor[rgb]{0,0,0}{\textbf{73\% positive}}
\end{flushright}

\end{minipage}};

\end{tikzpicture}
}

%% file: tables/attack_examples.tex
\begin{tabular}{ll}
\toprule

\multicolumn{2}{l}{\textbf{Original:} \negp{70\% Negative}} \\
\textbf{Input Example:} & in its best moments , resembles a bad high school production of grease , without benefit of song .   \\
\midrule

\multicolumn{2}{l}{\textbf{Genetic:} \posp{56\% Positive}} \\
\textbf{Adversarial Example:} & in its best \adv{moment} , \adv{recalling} a \adv{naughty} high school production of \adv{lubrication} , \adv{unless} benefit of song .   \\
\midrule

\multicolumn{2}{l}{\textbf{BAE:} \posp{56\% Positive}} \\
\textbf{Adversarial Example:} & in its best moments , resembles a \adv{great} high school production of grease , without benefit of song .   \\
\midrule

\multicolumn{2}{l}{\textbf{SO-Enum and SO-Beam (ours):} \negp{60\% Negative\;} \posp{(67\% Positive)}} \\
\textbf{Vulnerable Example:} & in its best moments , resembles a \negp{bad} \posp{(unhealthy)} high school production of \adv{musicals} , without benefit of song .   \\

\bottomrule
\end{tabular}

%% file: tables/attack_success_rate.tex
\begin{tabular}{lcc|cc}
\toprule
      & \multicolumn{4}{c}{\textbf{Attack Success Rate (\%)}}  \\
             &             Genetic & BAE & \makecell{SO-Enum} & \makecell{SO-Beam} \\
\midrule
\multicolumn{2}{l}{\textbf{Base Models:}} \\
         BoW &                   57.0 &  69.7 &    95.3 &    \first{99.7}  \\
         CNN &                   62.0 &  71.0 &    95.3 &    \first{99.8}  \\
        LSTM &                   60.0 &  68.3 &    95.8 &    \first{99.5}  \\
 Transformer &                   73.0 &  74.3 &    95.4 &    \first{98.0}  \\
   BERT-base &                   41.0 &  61.5 &    94.3 &    \first{98.7}  \\
\multicolumn{2}{l}{\textbf{Robust Models:}} \\
         BoW &                   28.0 &  63.1 &    81.5 &    \first{88.4}  \\
         CNN &                   23.0 &  64.4 &    91.0 &    \first{96.0}  \\
        LSTM &                   24.0 &  61.0 &    62.9 &    \first{77.5}  \\
 Transformer &                   56.0 &  71.6 &    91.2 &    \first{96.2}  \\
\bottomrule
\end{tabular}

%% file: tables/quality_gpt2_l0.tex
\begin{tabular}{lcccccc}
\toprule
             & \multicolumn{3}{c}{\textbf{SO-Enum}} & \multicolumn{3}{c}{\textbf{SO-Beam}} \\
      & \makecell{Original\\PPL} & \makecell{Perturb\\PPL} & \makecell{$\ell_0$} & \makecell{Original\\PPL} & \makecell{Perturb\\PPL} & \makecell{$\ell_0$} \\
\midrule
\multicolumn{2}{l}{\textbf{Base Models:}} \\
         BoW &      168 &       202 &    1.1  &      166 &       202   &     1.2 \\
         CNN &      170 &       204 &    1.1  &      166 &       201   &     1.2 \\
        LSTM &      168 &       204 &    1.1  &      166 &       204   &     1.2 \\
 Transformer &      165 &       193 &    1.0  &      165 &       195   &     1.1 \\
   BERT-base &      170 &       229 &    1.3  &      168 &       222   &     1.4 \\
\multicolumn{2}{l}{\textbf{Robust Models:}} \\
         BoW &      170 &       212 &    1.2  &      171 &       222   &     1.4 \\
         CNN &      166 &       209 &    1.2  &      168 &       210   &     1.3 \\
        LSTM &      194 &       251 &    1.3  &      185 &       260   &     1.8 \\
 Transformer &      170 &       213 &    1.2  &      165 &       208   &     1.3 \\
\bottomrule
\end{tabular}

%% file: tables/human_evaluation.tex
\begin{tabular}{cccc}
\toprule
\multicolumn{2}{c}{\textbf{Naturalness (1-5)}} & \multicolumn{2}{c}{\textbf{Semantic Similarity (\%)}} \\
\makecell{Original} & \makecell{Perturb} & \makecell{Original} & \makecell{Perturb} \\
\midrule
3.87 &       3.63 &    85 &       71 \\
\bottomrule
\end{tabular}

%% file: figures/bias.tex
\resizebox{.9\columnwidth}{!}{
\setlength\tabcolsep{1pt}
\begin{tabular}{rl}
\input{figures/biased-no} & \input{figures/biased-yes}
\end{tabular}
}


%% file: figures/biased-no.tex
  
\tikzset {_43cip8kdf/.code = {\pgfsetadditionalshadetransform{ \pgftransformshift{\pgfpoint{0 bp } { 0 bp }  }  \pgftransformrotate{-90 }  \pgftransformscale{2 }  }}}
\pgfdeclarehorizontalshading{_gk9siekrr}{150bp}{rgb(0bp)=(0.49,0.83,0.13);
rgb(41.517857142857146bp)=(0.49,0.83,0.13);
rgb(53.92857142857143bp)=(0.79,0.79,0.79);
rgb(100bp)=(0.79,0.79,0.79)}
\tikzset{every picture/.style={line width=0.75pt}} 

\begin{tikzpicture}[x=0.75pt,y=0.75pt,yscale=-1,xscale=1]
\clip (10,0.5) rectangle (200,200);


\draw  [draw opacity=0][shading=_gk9siekrr,_43cip8kdf] (0,0.5) -- (208.5,0.5) -- (208.5,200) -- (0,200) -- cycle ;

\draw  [fill={rgb, 255:red, 255; green, 247; blue, 115 }  ,fill opacity=1 ] (11.5,30.25) .. controls (11.5,25.14) and (15.64,21) .. (20.75,21) .. controls (25.86,21) and (30,25.14) .. (30,30.25) .. controls (30,35.36) and (25.86,39.5) .. (20.75,39.5) .. controls (15.64,39.5) and (11.5,35.36) .. (11.5,30.25) -- cycle ;
\draw  [fill={rgb, 255:red, 241; green, 89; blue, 108 }  ,fill opacity=1 ] (48.5,30.25) .. controls (48.5,25.14) and (52.64,21) .. (57.75,21) .. controls (62.86,21) and (67,25.14) .. (67,30.25) .. controls (67,35.36) and (62.86,39.5) .. (57.75,39.5) .. controls (52.64,39.5) and (48.5,35.36) .. (48.5,30.25) -- cycle ;
\draw    (30,30.25) -- (48.5,30.25) ;

\draw  [fill={rgb, 255:red, 255; green, 247; blue, 115 }  ,fill opacity=1 ] (27.05,115.38) .. controls (25.36,110.56) and (27.91,105.28) .. (32.74,103.6) .. controls (37.56,101.92) and (42.83,104.47) .. (44.52,109.29) .. controls (46.2,114.12) and (43.65,119.39) .. (38.82,121.07) .. controls (34,122.75) and (28.73,120.2) .. (27.05,115.38) -- cycle ;
\draw  [fill={rgb, 255:red, 241; green, 89; blue, 108 }  ,fill opacity=1 ] (61.98,103.21) .. controls (60.3,98.38) and (62.85,93.11) .. (67.68,91.43) .. controls (72.5,89.75) and (77.77,92.3) .. (79.45,97.12) .. controls (81.14,101.94) and (78.59,107.22) .. (73.76,108.9) .. controls (68.94,110.58) and (63.67,108.03) .. (61.98,103.21) -- cycle ;
\draw    (44.52,109.29) -- (61.98,103.21) ;

\draw  [fill={rgb, 255:red, 255; green, 247; blue, 115 }  ,fill opacity=1 ] (36.07,155.67) .. controls (37.09,150.67) and (41.98,147.44) .. (46.99,148.47) .. controls (51.99,149.5) and (55.22,154.39) .. (54.19,159.39) .. controls (53.16,164.39) and (48.27,167.62) .. (43.27,166.59) .. controls (38.26,165.56) and (35.04,160.67) .. (36.07,155.67) -- cycle ;
\draw  [fill={rgb, 255:red, 241; green, 89; blue, 108 }  ,fill opacity=1 ] (72.31,163.11) .. controls (73.34,158.11) and (78.23,154.88) .. (83.23,155.91) .. controls (88.24,156.94) and (91.46,161.83) .. (90.43,166.83) .. controls (89.41,171.83) and (84.52,175.06) .. (79.51,174.03) .. controls (74.51,173) and (71.28,168.11) .. (72.31,163.11) -- cycle ;
\draw    (54.19,159.39) -- (72.31,163.11) ;

\draw  [fill={rgb, 255:red, 255; green, 247; blue, 115 }  ,fill opacity=1 ] (136.4,178.35) .. controls (134.54,173.59) and (136.89,168.23) .. (141.65,166.37) .. controls (146.41,164.51) and (151.78,166.86) .. (153.63,171.62) .. controls (155.49,176.37) and (153.14,181.74) .. (148.38,183.6) .. controls (143.63,185.46) and (138.26,183.11) .. (136.4,178.35) -- cycle ;
\draw  [fill={rgb, 255:red, 241; green, 89; blue, 108 }  ,fill opacity=1 ] (170.87,164.88) .. controls (169.01,160.13) and (171.36,154.76) .. (176.12,152.9) .. controls (180.87,151.04) and (186.24,153.39) .. (188.1,158.15) .. controls (189.96,162.91) and (187.61,168.27) .. (182.85,170.13) .. controls (178.09,171.99) and (172.72,169.64) .. (170.87,164.88) -- cycle ;
\draw    (153.63,171.62) -- (170.87,164.88) ;

\draw  [fill={rgb, 255:red, 255; green, 247; blue, 115 }  ,fill opacity=1 ] (139.3,87.62) .. controls (140.53,82.66) and (145.54,79.62) .. (150.5,80.85) .. controls (155.46,82.07) and (158.49,87.08) .. (157.27,92.04) .. controls (156.05,97) and (151.04,100.03) .. (146.08,98.81) .. controls (141.11,97.59) and (138.08,92.58) .. (139.3,87.62) -- cycle ;
\draw  [fill={rgb, 255:red, 241; green, 89; blue, 108 }  ,fill opacity=1 ] (175.23,96.46) .. controls (176.45,91.5) and (181.46,88.47) .. (186.42,89.69) .. controls (191.39,90.91) and (194.42,95.92) .. (193.2,100.88) .. controls (191.97,105.84) and (186.96,108.88) .. (182,107.65) .. controls (177.04,106.43) and (174.01,101.42) .. (175.23,96.46) -- cycle ;
\draw    (157.27,92.04) -- (175.23,96.46) ;

\draw (14,7) node [anchor=north west][inner sep=0.75pt]    {\Large $x$};
\draw (49,5) node [anchor=north west][inner sep=0.75pt]    {\Large $x \oplus \p$};

\end{tikzpicture}

%% file: figures/biased-yes.tex
  
\tikzset {_t2qv7a4xh/.code = {\pgfsetadditionalshadetransform{ \pgftransformshift{\pgfpoint{0 bp } { 0 bp }  }  \pgftransformrotate{0 }  \pgftransformscale{2 }  }}}
\pgfdeclarehorizontalshading{_8vy4xm3ip}{150bp}{rgb(0bp)=(0.49,0.83,0.13);
rgb(37.5bp)=(0.49,0.83,0.13);
rgb(52.589285714285715bp)=(0.79,0.79,0.79);
rgb(100bp)=(0.79,0.79,0.79)}
\tikzset{every picture/.style={line width=0.75pt}} 

\begin{tikzpicture}[x=0.75pt,y=0.75pt,yscale=-1,xscale=1]
\clip (10,0.5) rectangle (200,200);

\draw  [draw opacity=0][shading=_8vy4xm3ip,_t2qv7a4xh] (0,0.5) -- (208.5,0.5) -- (208.5,270) -- (0,270) -- cycle ;
\draw  [fill={rgb, 255:red, 255; green, 247; blue, 115 }  ,fill opacity=1 ] (11.5,30.25) .. controls (11.5,25.14) and (15.64,21) .. (20.75,21) .. controls (25.86,21) and (30,25.14) .. (30,30.25) .. controls (30,35.36) and (25.86,39.5) .. (20.75,39.5) .. controls (15.64,39.5) and (11.5,35.36) .. (11.5,30.25) -- cycle ;
\draw  [fill={rgb, 255:red, 241; green, 89; blue, 108 }  ,fill opacity=1 ] (48.5,30.25) .. controls (48.5,25.14) and (52.64,21) .. (57.75,21) .. controls (62.86,21) and (67,25.14) .. (67,30.25) .. controls (67,35.36) and (62.86,39.5) .. (57.75,39.5) .. controls (52.64,39.5) and (48.5,35.36) .. (48.5,30.25) -- cycle ;
\draw    (30,30.25) -- (48.5,30.25) ;

\draw  [fill={rgb, 255:red, 255; green, 247; blue, 115 }  ,fill opacity=1 ] (27.05,115.38) .. controls (25.36,110.56) and (27.91,105.28) .. (32.74,103.6) .. controls (37.56,101.92) and (42.83,104.47) .. (44.52,109.29) .. controls (46.2,114.12) and (43.65,119.39) .. (38.82,121.07) .. controls (34,122.75) and (28.73,120.2) .. (27.05,115.38) -- cycle ;
\draw  [fill={rgb, 255:red, 241; green, 89; blue, 108 }  ,fill opacity=1 ] (61.98,103.21) .. controls (60.3,98.38) and (62.85,93.11) .. (67.68,91.43) .. controls (72.5,89.75) and (77.77,92.3) .. (79.45,97.12) .. controls (81.14,101.94) and (78.59,107.22) .. (73.76,108.9) .. controls (68.94,110.58) and (63.67,108.03) .. (61.98,103.21) -- cycle ;
\draw    (44.52,109.29) -- (61.98,103.21) ;

\draw  [fill={rgb, 255:red, 255; green, 247; blue, 115 }  ,fill opacity=1 ] (36.07,155.67) .. controls (37.09,150.67) and (41.98,147.44) .. (46.99,148.47) .. controls (51.99,149.5) and (55.22,154.39) .. (54.19,159.39) .. controls (53.16,164.39) and (48.27,167.62) .. (43.27,166.59) .. controls (38.26,165.56) and (35.04,160.67) .. (36.07,155.67) -- cycle ;
\draw  [fill={rgb, 255:red, 241; green, 89; blue, 108 }  ,fill opacity=1 ] (72.31,163.11) .. controls (73.34,158.11) and (78.23,154.88) .. (83.23,155.91) .. controls (88.24,156.94) and (91.46,161.83) .. (90.43,166.83) .. controls (89.41,171.83) and (84.52,175.06) .. (79.51,174.03) .. controls (74.51,173) and (71.28,168.11) .. (72.31,163.11) -- cycle ;
\draw    (54.19,159.39) -- (72.31,163.11) ;

\draw  [fill={rgb, 255:red, 255; green, 247; blue, 115 }  ,fill opacity=1 ] (136.4,178.35) .. controls (134.54,173.59) and (136.89,168.23) .. (141.65,166.37) .. controls (146.41,164.51) and (151.78,166.86) .. (153.63,171.62) .. controls (155.49,176.37) and (153.14,181.74) .. (148.38,183.6) .. controls (143.63,185.46) and (138.26,183.11) .. (136.4,178.35) -- cycle ;
\draw  [fill={rgb, 255:red, 241; green, 89; blue, 108 }  ,fill opacity=1 ] (170.87,164.88) .. controls (169.01,160.13) and (171.36,154.76) .. (176.12,152.9) .. controls (180.87,151.04) and (186.24,153.39) .. (188.1,158.15) .. controls (189.96,162.91) and (187.61,168.27) .. (182.85,170.13) .. controls (178.09,171.99) and (172.72,169.64) .. (170.87,164.88) -- cycle ;
\draw    (153.63,171.62) -- (170.87,164.88) ;

\draw  [fill={rgb, 255:red, 255; green, 247; blue, 115 }  ,fill opacity=1 ] (139.3,87.62) .. controls (140.53,82.66) and (145.54,79.62) .. (150.5,80.85) .. controls (155.46,82.07) and (158.49,87.08) .. (157.27,92.04) .. controls (156.05,97) and (151.04,100.03) .. (146.08,98.81) .. controls (141.11,97.59) and (138.08,92.58) .. (139.3,87.62) -- cycle ;
\draw  [fill={rgb, 255:red, 241; green, 89; blue, 108 }  ,fill opacity=1 ] (175.23,96.46) .. controls (176.45,91.5) and (181.46,88.47) .. (186.42,89.69) .. controls (191.39,90.91) and (194.42,95.92) .. (193.2,100.88) .. controls (191.97,105.84) and (186.96,108.88) .. (182,107.65) .. controls (177.04,106.43) and (174.01,101.42) .. (175.23,96.46) -- cycle ;
\draw    (157.27,92.04) -- (175.23,96.46) ;

\draw (14,7) node [anchor=north west][inner sep=0.75pt]    {\Large $x$};
\draw (49,5) node [anchor=north west][inner sep=0.75pt]    {\Large $x \oplus \p$};

\end{tikzpicture}

%% file: tables/filtered_stats.tex
\begin{tabular}{lrr}
\toprule
Patch Words &  \# Original & \# Perturbed \\
\midrule
    he,she    &    \num{5} &   \num{325401} \\
   his,her    &    \num{4} &   \num{255245} \\
   him,her    &    \num{4} &   \num{233803} \\
 men,women    &    \num{3} &   \num{192504} \\
 man,woman    &    \num{3} &   \num{222981} \\
actor,actress &    \num{2} &   \num{141780} \\
 $\dots$ &     &    \\
\midrule
 Total &   \num{34} &  \num{2317635} \\
\bottomrule
\end{tabular}

%% file: figures/ctb-per-token.tex
\resizebox{\columnwidth}{!}{
\begin{tabular}{c}
\includegraphics{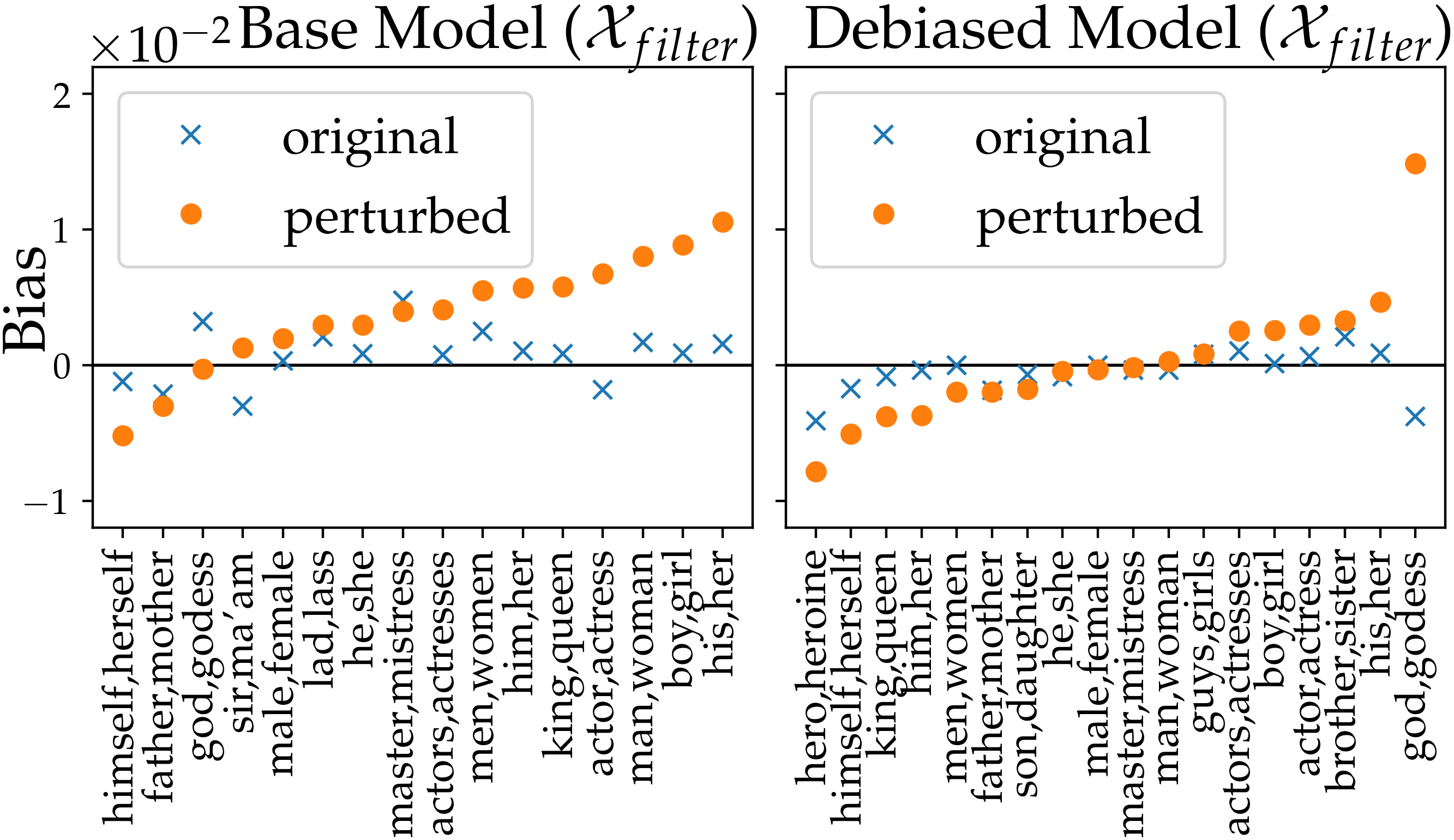}
\end{tabular}
}

%% file: figures/hidden-bias.tex
\includegraphics[width=.8\textwidth]{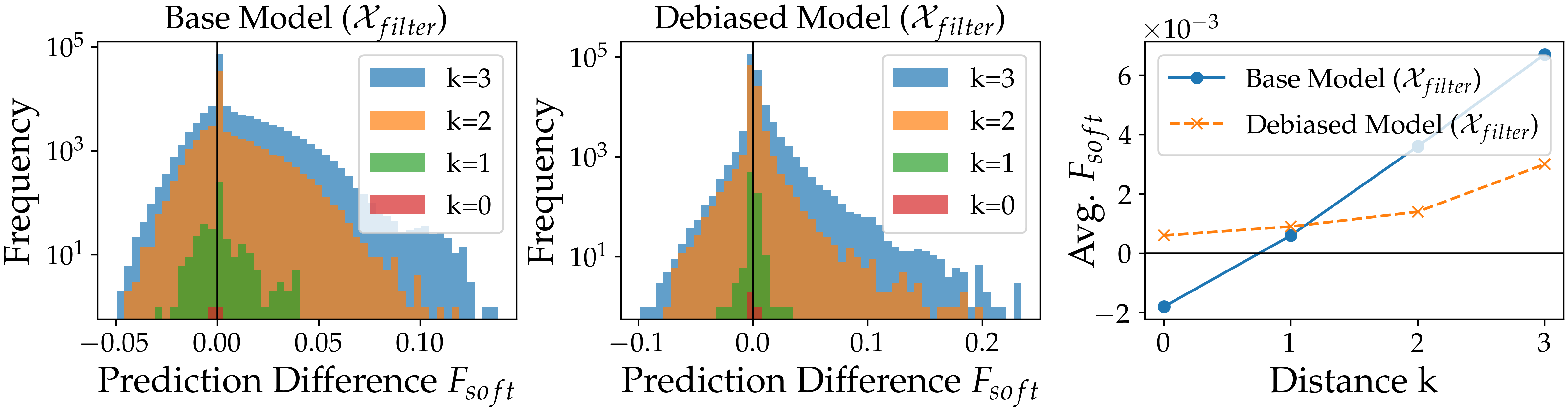}

%% file: appendix.tex




\section{Supplemental Material}

\subsection{Random Baseline}
\label{apd:random-baseline}
To validate the effectiveness of minimizing \cref{eq:loss}, we also experiment on a \so baseline that constructs vulnerable examples by randomly replacing up to 6 words. We use the same masked language model and threshold as \sob such that they share a similar neighborhood. We perform the attack on the same models as \cref{tb:attack_success}, and the attack success rates on robustly trained BoW, CNN, LSTM, and Transformers are 18.8\%, 22.3\%, 15.2\%, and 25.1\%, respectively.
Despite being a \so attack, the random baseline has low attack success rates thus demonstrates the effectiveness of \sob.

\subsection{Human Evaluation}
\label{apd:human}

We randomly select 100 successful attacks from \sob and consider four types of examples (for a total of 400 examples): The original examples with and without synonym substitution $\p$, and the vulnerable examples with and without synonym substitution $\p$. For each example, we annotate the naturalness and sentiment separately as described below.

\define{Naturalness of vulnerable examples.} We ask the annotators to score the likelihood of being an original example (i.e., not altered by computer) based on grammar correctness and naturalness, with a Likert scale of 1-5: (1) Sure adversarial example. (2) Likely an adversarial example. (3) Neutral. (4) Likely an original example. (5) Sure original example.

\define{Semantic similarity after the synonym substitution.} We first ask the annotators to predict the sentiment on a Likert scale of 1-5, and then map the prediction to three categories: negative, neutral, and positive. We consider two examples to have the same semantic meaning if and only if they are both positive or negative.

\subsection{Running Time}
\label{apd:runtime}

We experiment with the validation split on a single RTX 3090, and measure the average running time per example. As shown in \cref{tb:runtime}, \sob runs faster than \soe since it utilizes the probability output. The running time may increase if the model has improved \so robustness.

\begin{table}[!h]
    \centering
    \resizebox{.9\columnwidth}{!}{\input{tables/runtime}}
    \caption{The average running time over 872 examples (100 for Genetic due to long running time).}
    \label{tb:runtime}
\end{table}

\subsection{Additional Results on Protected Tokens}
\label{apd:additional-tokens}

\begin{figure}[h]
    \centering
    \includegraphics[width=\columnwidth]{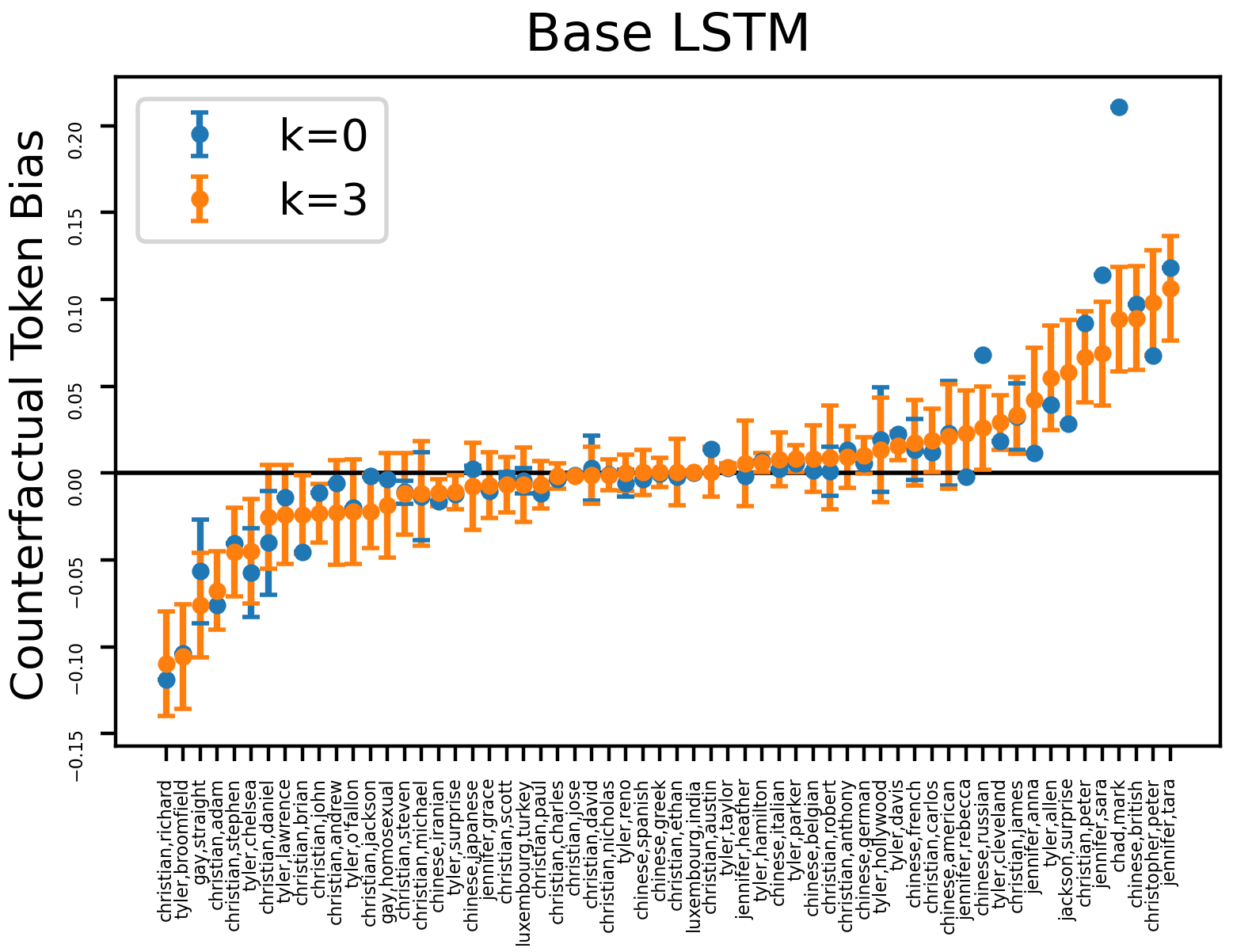}
    \caption{Additional counterfactual token bias measured on the original validation split with base LSTM.}
    \label{fig:checklist-bias}
\end{figure}

\cref{fig:checklist-bias} presents the experimental results with additional protected tokens such as nationality, religion, and sexual orientation (from \citet{Marco2020checklist}). We use the same base LSTM as described in \cref{sec:exp-bias}. One interesting observation is when $p  = (\w{gay}, \w{straight})$ where the bias is negative, indicating that the sentiment classifier tends to give more \emph{negative} prediction when substituting $\w{gay} \rightarrow \w{straight}$ in the input.
This phenomenon is opposite to the behavior of toxicity classifiers \citep{dixon2018measuring}, and we hypothesize that it may be caused by the different distribution of training data. To verify the hypothesis, we count the number of training examples containing each word, and observe that we have far more negative examples than positive examples among those containing \w{straight} (\cref{tb:gay-train-stats}). After looking into the training set, it turns out that \w{straight to video} is a common phrase to criticize a film, thus the classifier incorrectly correlates \w{straight} with negative sentiment. This also reveals the limitation of our method on polysemous words.

\begin{table}[h]
    \centering
    \resizebox{.7\columnwidth}{!}{\input{tables/gay-train-stats}}
    \caption{Number of negative and positive examples containing \w{gay} and \w{straight} in the training set.}
    \label{tb:gay-train-stats}
\end{table}

\begin{figure}[h]
    \centering
    \includegraphics[width=\columnwidth]{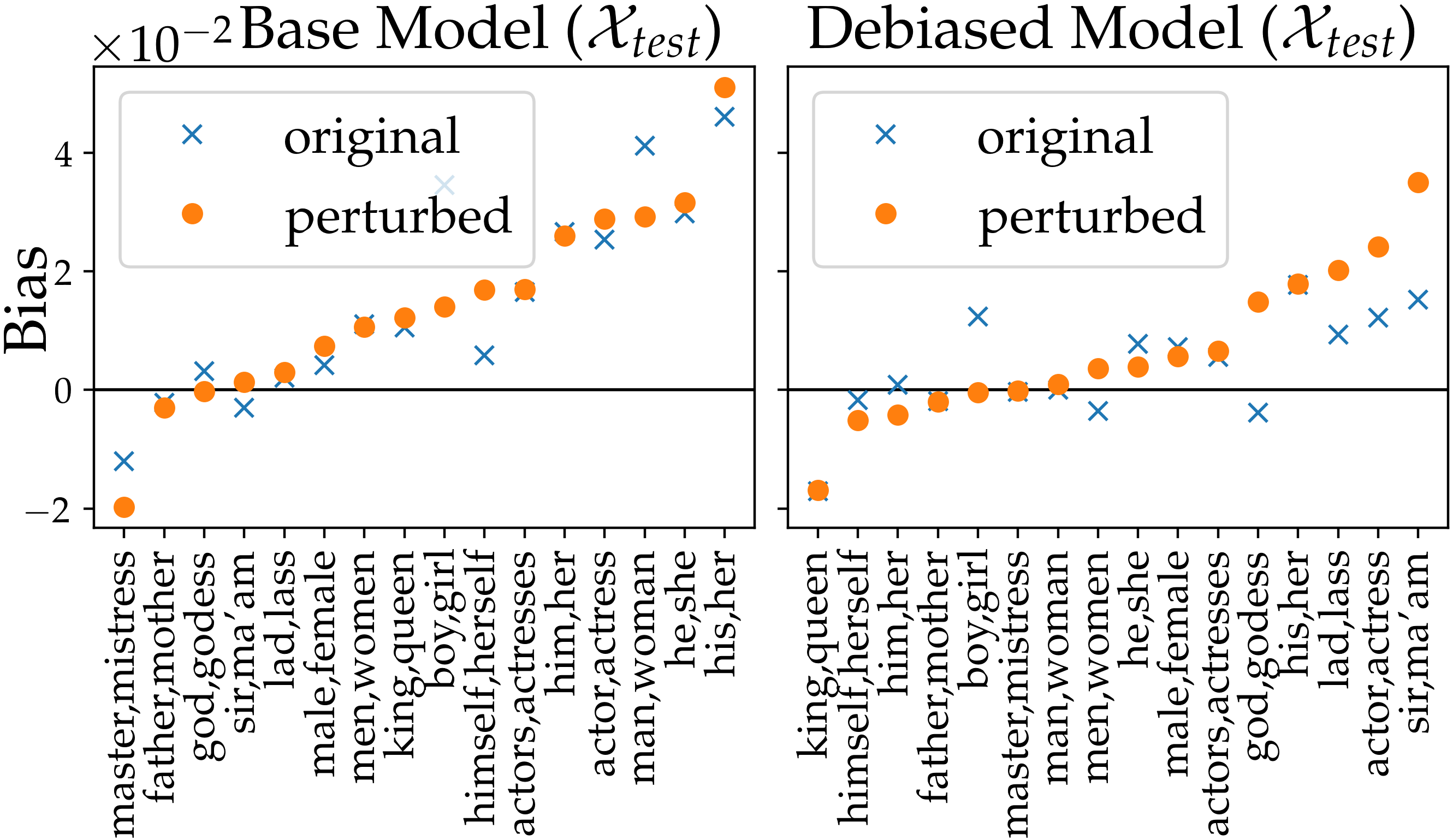}
    \caption{Full results for gendered tokens measured on the original validation split.}
    \label{fig:full-gender}
\end{figure}

In \cref{fig:full-gender}, we measure the bias on $\cX_\text{test}$ and observe positive bias on most tokens for both $k=0$ and $k=3$, which indicates that the model \qt{tends} to make more positive predictions for examples containing certain female pronouns than male pronouns. Notice that even though gender swap mitigates the bias to some extent, it is still difficult to fully eliminate the bias. This is probably caused by tuples like (\w{him}, \w{his}, \w{her}) which cannot be swapped perfectly, and requires additional processing such as part-of-speech resolving~\citep{zhao-etal-2018-gender}.

To help evaluate the naturalness of our constructed examples used in \cref{sec:bias}, we provide sample sentences in \cref{tb:bias_examples_lstm} and  \cref{tb:bias_examples2_lstm}.
Bold words are the corresponding patch words $\p$, taken from the predefined list of gendered pronouns.

\subsection{Additional Results on Robustness}
\label{apd:additional-robustness}

\cref{tb:fo_quality} provides the quality metrics for \fo attacks, where we measure the GPT-2 perplexity and $\ell_0$ norm distance between the input and the adversarial example. For BAE we evaluate on 872 validation examples, and for Genetic we evaluate on 100 validation examples due to the long running time.

\begin{table}[h]
    \centering
    \resizebox{.9\columnwidth}{!}{\input{tables/fo_quality_gpt2_l0}}
    \caption{The quality metrics for \fo attacks from successful attacks. We compare median perplexities (PPL) and average $\ell_0$ norm distances.}
    \label{tb:fo_quality}
\end{table}

\cref{tb:additional_sobeam_lstm_normal} shows additional attack results from SO-Beam on base LSTM, and \cref{tb:additional_sobeam_cnn_cert} shows additional attack results from SO-Beam on robust CNN. Bold words are the corresponding patch words $\p$, taken from the predefined list of counter-fitted synonyms.

\begin{table*}[h]
    \centering
    \resizebox{\textwidth}{!}{\input{tables/bias_examples}}
    \caption{Additional counterfactual bias examples on base LSTM with $\p = (\w{actor}, \w{actress})$. We only present 5 examples per $k$ due to space constrain.}
    \label{tb:bias_examples_lstm}
\end{table*}

\begin{table*}[h]
    \centering
    \resizebox{\textwidth}{!}{\input{tables/bias_examples2}}
    \caption{Additional counterfactual bias examples on base LSTM with $\p = (\w{boy}, \w{girl})$. We only present 5 examples per $k$ due to space constrain.}
    \label{tb:bias_examples2_lstm}
\end{table*}

\begin{table*}[h]
    \centering
    \resizebox{\textwidth}{!}{\input{tables/additional_sobeam_lstm_normal}}
    \caption{Additional sentiment classification results from SO-Beam on base LSTM.}
    \label{tb:additional_sobeam_lstm_normal}
\end{table*}

\clearpage

\begin{table*}[h]
    \centering
    \resizebox{\textwidth}{!}{\input{tables/additional_sobeam_cnn_cert}}
    \caption{Additional sentiment classification results from SO-Beam on robust CNN.}
    \label{tb:additional_sobeam_cnn_cert}
\end{table*}

%% file: tables/runtime.tex
\begin{tabular}{lcc|cc}
\toprule
      & \multicolumn{4}{c}{\textbf{Running Time (seconds)}}  \\
             &             Genetic & BAE & \makecell{SO-Enum} & \makecell{SO-Beam} \\
\midrule
\multicolumn{2}{l}{\textbf{Base Models:}} \\
         BoW &                   31.6 &           0.9 &      6.2 &             1.8 \\
         CNN &                   28.8 &           1.0 &      5.9 &             1.7 \\
        LSTM &                   31.9 &           1.1 &      7.0 &             1.9 \\
 Transformer &                   51.9 &           0.5 &      6.5 &             2.5 \\
   BERT-base &                   65.6 &           1.1 &     35.4 &             7.1 \\
\multicolumn{2}{l}{\textbf{Robust Models:}} \\
         BoW &                   103.9 &           1.0 &      8.0 &             3.5 \\
         CNN &                   129.4 &           1.0 &      6.7 &             2.6 \\
        LSTM &                   116.4 &           1.1 &     10.7 &             5.3 \\
 Transformer &                   66.4 &           0.5 &      5.9 &             2.6 \\
\bottomrule
\end{tabular}

%% file: tables/gay-train-stats.tex
\begin{tabular}{lcc}
\toprule
 & \# Negative & \# Positive \\
\midrule
\w{gay} & 37 & 20 \\
\w{straight} & 71 & 18 \\
\bottomrule
\end{tabular}

%% file: tables/fo_quality_gpt2_l0.tex
\begin{tabular}{lcccccc}
\toprule
             & \multicolumn{3}{c}{\textbf{Genetic}} & \multicolumn{3}{c}{\textbf{BAE}} \\
      & \makecell{Original\\PPL} & \makecell{Perturb\\PPL} & \makecell{$\ell_0$} & \makecell{Original\\PPL} & \makecell{Perturb\\PPL} & \makecell{$\ell_0$} \\
\midrule
\multicolumn{2}{l}{\textbf{Base Models:}} \\
         BoW &      145 &       258 &    3.3  &      192 &       268 &     1.6 \\
         CNN &      146 &       282 &    3.0  &      186 &       254 &     1.5 \\
        LSTM &      131 &       238 &    2.9  &      190 &       263 &     1.6 \\
 Transformer &      137 &       232 &    2.8  &      185 &       254 &     1.4 \\
   BERT-base &      201 &       342 &    3.4  &      189 &       277 &     1.6 \\
\multicolumn{2}{l}{\textbf{Robust Models:}} \\
         BoW &      132 &       177 &    2.4  &      214 &       269 &     1.5 \\
         CNN &      136 &       236 &    2.7  &      211 &       279 &     1.5 \\
        LSTM &      163 &       267 &    2.5  &      220 &       302 &     1.6 \\
 Transformer &      118 &       200 &    2.8  &      196 &       261 &     1.4 \\
\bottomrule
\end{tabular}

%% file: tables/bias_examples.tex
\begin{tabularx}{1.23\textwidth}{ccX}
\toprule
\multicolumn{1}{c}{\textbf{Type}} & \multicolumn{1}{c}{\textbf{Predictions}} & \multicolumn{1}{c}{\textbf{Text}} \\
\midrule

Original 
& \negp{95\% Negative\;} \negp{94\% Negative\;} & it 's hampered by a lifetime-channel kind of plot and a lead \textbf{actor (actress)} who is out of their depth . \\

\midrule
Distance $k=1$
& \negp{97\% Negative} \negp{(97\% Negative)} & it 's hampered by a lifetime-channel kind of plot and lone lead \textbf{actor (actress)} who is out of their depth . \\
& \negp{56\% Negative} \posp{(55\% Positive\;)} & it 's hampered by a lifetime-channel kind of plot and a lead \textbf{actor (actress)} who is out of creative depth . \\
& \negp{89\% Negative} \negp{(84\% Negative)} & it 's hampered by a lifetime-channel kind of plot and a lead \textbf{actor (actress)} who talks out of their depth . \\
& \negp{98\% Negative} \negp{(98\% Negative)} & it 's hampered by a lifetime-channel kind of plot and a lead \textbf{actor (actress)} who is out of production depth . \\
& \negp{96\% Negative} \negp{(96\% Negative)} & it 's hampered by a lifetime-channel kind of plot and a lead \textbf{actor (actress)} that is out of their depth . \\

\midrule
Distance $k=2$
& \negp{88\% Negative} \negp{(87\% Negative)} & it 's hampered by a lifetime-channel cast of stars and a lead \textbf{actor (actress)} who is out of their depth . \\
& \negp{96\% Negative} \negp{(95\% Negative)} & it 's hampered by a simple set of plot and a lead \textbf{actor (actress)} who is out of their depth . \\
& \negp{54\% Negative} \negp{(54\% Negative)} & it 's framed about a lifetime-channel kind of plot and a lead \textbf{actor (actress)} who is out of their depth . \\
& \negp{90\% Negative} \negp{(88\% Negative)} & it 's hampered by a lifetime-channel mix between plot and a lead \textbf{actor (actress)} who is out of their depth . \\
& \negp{78\% Negative} \negp{(68\% Negative)} & it 's hampered by a lifetime-channel kind of plot and a lead \textbf{actor (actress)} who storms out of their mind . \\

\midrule
Distance $k=3$
& \posp{52\% Positive\;} \posp{(64\% Positive\;)} & it 's characterized by a lifetime-channel combination comedy plot and a lead \textbf{actor (actress)} who is out of their depth . \\
& \negp{93\% Negative} \negp{(93\% Negative)} & it 's hampered by a lifetime-channel kind of star and a lead \textbf{actor (actress)} who falls out of their depth . \\
& \negp{58\% Negative} \negp{(57\% Negative)} & it 's hampered by a tough kind of singer and a lead \textbf{actor (actress)} who is out of their teens . \\
& \negp{70\% Negative} \negp{(52\% Negative)} & it 's hampered with a lifetime-channel kind of plot and a lead \textbf{actor (actress)} who operates regardless of their depth . \\
& \negp{58\% Negative} \posp{(53\% Positive\;)} & it 's hampered with a lifetime-channel cast of plot and a lead \textbf{actor (actress)} who is out of creative depth . \\

\bottomrule
\end{tabularx}

%% file: tables/bias_examples2.tex
\begin{tabularx}{1.23\textwidth}{ccX}
\toprule
\multicolumn{1}{c}{\textbf{Type}} & \multicolumn{1}{c}{\textbf{Predictions}} & \multicolumn{1}{c}{\textbf{Text}} \\
\midrule

Original
& \posp{55\% Positive\;} \posp{(67\% Positive\;)} & a hamfisted romantic comedy that makes our \textbf{boy (girl)} the hapless facilitator of an extended cheap shot across the mason-dixon line . \\

\midrule
Distance $k=1$
& \posp{52\% Positive\;} \posp{(66\% Positive\;)} & a hamfisted romantic comedy that makes our \textbf{boy (girl)} the hapless facilitator of an extended cheap shot from the mason-dixon line . \\
& \posp{73\% Positive\;} \posp{(79\% Positive\;)} & a hamfisted romantic comedy that makes our \textbf{boy (girl)} the hapless facilitator gives an extended cheap shot across the mason-dixon line . \\
& \negp{56\% Negative} \posp{(58\% Positive\;)} & a hamfisted romantic comedy that makes our \textbf{boy (girl)} the hapless facilitator of an extended cheap shot across the phone line . \\
& \posp{75\% Positive\;} \posp{(83\% Positive\;)} & a hamfisted romantic comedy that makes our \textbf{boy (girl)} the hapless facilitator of an extended chase shot across the mason-dixon line . \\
& \posp{75\% Positive\;} \posp{(81\% Positive\;)} & a hamfisted romantic comedy that makes our \textbf{boy (girl)} our hapless facilitator of an extended cheap shot across the mason-dixon line . \\

\midrule
Distance $k=2$
& \posp{85\% Positive\;} \posp{(85\% Positive\;)} & a hilarious romantic comedy that makes our \textbf{boy (girl)} the hapless facilitator of an emotionally cheap shot across the mason-dixon line . \\
& \posp{81\% Positive\;} \posp{(86\% Positive\;)} & a hamfisted romantic comedy romance makes our \textbf{boy (girl)} the hapless facilitator of an extended cheap delivery across the mason-dixon line . \\
& \posp{84\% Positive\;} \posp{(87\% Positive\;)} & a hamfisted romantic romance adventure makes our \textbf{boy (girl)} the hapless facilitator of an extended cheap shot across the mason-dixon line . \\
& \negp{50\% Negative} \posp{(62\% Positive\;)} & a hamfisted romantic comedy that makes our \textbf{boy (girl)} the hapless boss of an extended cheap shot behind the mason-dixon line . \\
& \negp{77\% Negative} \negp{(71\% Negative)} & a hamfisted lesbian comedy that makes our \textbf{boy (girl)} the hapless facilitator of an extended slap shot across the mason-dixon line . \\

\midrule
Distance $k=3$
& \posp{97\% Positive\;} \posp{(97\% Positive\;)} & a darkly romantic comedy romance makes our \textbf{boy (girl)} the hapless facilitator delivers an extended cheap shot across the mason-dixon line . \\
& \posp{69\% Positive\;} \posp{(74\% Positive\;)} & a hamfisted romantic comedy film makes our \textbf{boy (girl)} the hapless facilitator of an extended cheap shot across the production line . \\
& \posp{87\% Positive\;} \posp{(89\% Positive\;)} & a hamfisted romantic comedy that makes our \textbf{boy (girl)} the exclusive focus of an extended cheap shot across the mason-dixon line . \\
& \posp{64\% Positive\;} \posp{(76\% Positive\;)} & a hamfisted romantic comedy that makes our \textbf{boy (girl)} the hapless facilitator shoots an extended flash shot across the camera line . \\
& \posp{99\% Positive\;} \posp{(99\% Positive\;)} & a compelling romantic comedy that makes our \textbf{boy (girl)} the perfect facilitator of an extended story shot across the mason-dixon line . \\

\bottomrule
\end{tabularx}

%% file: tables/additional_sobeam_lstm_normal.tex
\begin{tabularx}{1.23\textwidth}{lcX}
\toprule
\multicolumn{1}{c}{\textbf{Type}}  & \multicolumn{1}{c}{\textbf{Predictions}} & \multicolumn{1}{c}{\textbf{Text}} \\
\midrule

Original & \posp{99\% Positive\;} \posp{(99\% Positive\;)} & it 's a charming and \textbf{sometimes (often)} affecting journey .   \\
Vulnerable & \negp{59\% Negative} \posp{(56\% Positive\;)} & it 's a charming and \textbf{sometimes (often)} \adv{painful} journey .   \\
\midrule
Original & \negp{99\% Negative} \negp{(97\% Negative)} & unflinchingly \textbf{bleak (somber)} and desperate   \\
Vulnerable & \negp{80\% Negative} \posp{(79\% Positive\;)} & unflinchingly \textbf{bleak (somber)} and \adv{mysterious}   \\
\midrule
Original & \posp{99\% Positive\;} \posp{(93\% Positive\;)} & allows us to hope that nolan is poised to embark a major \textbf{career (quarry)} as a commercial yet inventive filmmaker .   \\
Vulnerable & \posp{76\% Positive\;} \negp{(75\% Negative)} & allows us to hope that nolan is poised to embark a major \textbf{career (quarry)} as a commercial yet \adv{amateur} filmmaker .   \\
\midrule
Original & \posp{94\% Positive\;} \posp{(68\% Positive\;)} & the acting , costumes , music , cinematography and sound are all \textbf{astounding (staggering)} given the production 's austere locales .   \\
Vulnerable & \posp{87\% Positive\;} \negp{(66\% Negative)} & the acting , costumes , music , cinematography and sound are \adv{largely} \textbf{astounding (staggering)} given the production 's austere locales .   \\
\midrule
Original & \posp{99\% Positive\;} \posp{(97\% Positive\;)} & although laced with humor and a few fanciful touches , the film is a refreshingly serious look at \textbf{young (juvenile)} women .   \\
Vulnerable & \posp{94\% Positive\;} \negp{(81\% Negative)} & although laced with humor and a few fanciful touches , the film is a \adv{moderately} serious look at \textbf{young (juvenile)} women .   \\
\midrule
Original & \negp{99\% Negative} \negp{(98\% Negative)} & a \textbf{sometimes (occasionally)} tedious film .   \\
Vulnerable & \negp{62\% Negative} \posp{(55\% Positive\;)} & a \textbf{sometimes (occasionally)} \adv{disturbing} film .   \\
\midrule
Original & \negp{100\% Negative} \negp{(100\% Negative)} & in exactly 89 minutes , most of which passed as slowly as if i 'd been sitting naked on an igloo , formula 51 sank from \textbf{quirky (lunatic)} to jerky to utter turkey .   \\
Vulnerable & \posp{51\% Positive\;} \negp{(65\% Negative)} & \adv{lasting} exactly 89 minutes , most of which passed as slowly as if i 'd been sitting naked on an igloo , \adv{but} 51 \adv{ranges} from \textbf{quirky (lunatic)} to \adv{delicious} to \adv{crisp} turkey .   \\
\midrule
Original & \posp{97\% Positive\;} \posp{(100\% Positive\;)} & the \textbf{scintillating (mesmerizing)} performances of the leads keep the film grounded and keep the audience riveted .   \\
Vulnerable & \negp{91\% Negative} \posp{(90\% Positive\;)} & the \textbf{scintillating (mesmerizing)} performances of the leads keep the film grounded and keep the \adv{plot} \adv{predictable} .   \\
\midrule
Original & \negp{89\% Negative} \negp{(96\% Negative)} & it takes a \textbf{uncanny (strange)} kind of laziness to waste the talents of robert forster , anne meara , eugene levy , and reginald veljohnson all in the same movie .   \\
Vulnerable & \posp{80\% Positive\;} \negp{(76\% Negative)} & it takes a \textbf{uncanny (strange)} kind of \adv{humour} to waste the talents of robert forster , anne meara , eugene levy , and reginald veljohnson all in the same movie .   \\
\midrule
Original & \negp{100\% Negative} \negp{(100\% Negative)} & ... the film suffers from a lack of humor ( something needed to \textbf{balance (equilibrium)} out the violence ) ...   \\
Vulnerable & \posp{76\% Positive\;} \negp{(86\% Negative)} & ... the film \adv{derives} from a \adv{lot} of humor ( something \adv{clever} to \textbf{balance (equilibrium)} out the violence ) ...   \\
\midrule
Original & \posp{55\% Positive\;} \posp{(97\% Positive\;)} & we root for ( clara and paul ) , even like them , though perhaps it 's an emotion closer to \textbf{pity (compassion)} .   \\
Vulnerable & \negp{89\% Negative} \posp{(91\% Positive\;)} & we root for ( clara and paul ) , even like them , though perhaps it 's an \adv{explanation} closer to \textbf{pity (compassion)} .   \\
\midrule
Original & \negp{95\% Negative} \negp{(97\% Negative)} & even horror \textbf{fans (stalkers)} will most likely not find what they 're seeking with trouble every day ; the movie lacks both thrills and humor .   \\
Vulnerable & \posp{61\% Positive\;} \negp{(59\% Negative)} & even horror \textbf{fans (stalkers)} will most likely not find what they 're seeking with trouble every day ; the movie \adv{has} both thrills and humor .   \\
\midrule
Original & \posp{100\% Positive\;} \posp{(100\% Positive\;)} & a gorgeous , high-spirited musical from india that exquisitely \textbf{mixed (blends)} music , dance , song , and high drama .   \\
Vulnerable & \negp{87\% Negative} \posp{(81\% Positive\;)} & a \adv{dark} , high-spirited musical from \adv{nowhere} that \adv{loosely} \textbf{mixed (blends)} music , dance , song , and high drama .   \\
\midrule
Original & \negp{99\% Negative} \negp{(94\% Negative)} & ... the movie is just a plain \textbf{old (longtime)} monster .   \\
Vulnerable & \negp{94\% Negative} \posp{(94\% Positive\;)} & ... the movie is just a \adv{pretty} \textbf{old (longtime)} monster .   \\

\bottomrule
\end{tabularx}

%% file: tables/additional_sobeam_cnn_cert.tex
\begin{tabularx}{1.23\textwidth}{lcX}
\toprule
\multicolumn{1}{c}{\textbf{Type}}  & \multicolumn{1}{c}{\textbf{Predictions}} & \multicolumn{1}{c}{\textbf{Text}} \\
\midrule

Original & \posp{54\% Positive\;} \posp{(69\% Positive\;)} & for the most part , director anne-sophie birot 's first feature is a sensitive , \textbf{overly (extraordinarily)} well-acted drama .   \\
Vulnerable & \negp{53\% Negative} \posp{(62\% Positive\;)} & for the most part , director anne-sophie \adv{benoit} 's first feature is a sensitive , \textbf{overly (extraordinarily)} well-acted drama .   \\
\midrule
Original & \posp{66\% Positive\;} \posp{(72\% Positive\;)} & mr. tsai is a very original \textbf{painter (artist)} in his medium , and what time is it there ?   \\
Vulnerable & \negp{52\% Negative} \posp{(55\% Positive\;)} & mr. tsai is a very original \textbf{painter (artist)} in his medium , and what time \adv{was} it there ?   \\
\midrule
Original & \posp{80\% Positive\;} \posp{(64\% Positive\;)} & sade is an \textbf{engaging (engage)} look at the controversial eponymous and fiercely atheistic hero .   \\
Vulnerable & \posp{53\% Positive\;} \negp{(66\% Negative)} & sade is an \textbf{engaging (engage)} look at the controversial eponymous \adv{or} fiercely atheistic hero .   \\
\midrule
Original & \negp{50\% Negative} \negp{(57\% Negative)} & so devoid of any kind of \textbf{comprehensible (intelligible)} story that it makes films like xxx and collateral damage seem like thoughtful treatises   \\
Vulnerable & \posp{53\% Positive\;} \negp{(54\% Negative)} & so devoid of any kind of \textbf{comprehensible (intelligible)} story that it makes films like xxx and collateral \adv{2} seem like thoughtful treatises   \\
\midrule
Original & \posp{90\% Positive\;} \posp{(87\% Positive\;)} & a tender , \textbf{heartfelt (deepest)} family drama .   \\
Vulnerable & \posp{60\% Positive\;} \negp{(61\% Negative)} & a \adv{somber} , \textbf{heartfelt (deepest)} \adv{funeral} drama .   \\
\midrule
Original & \posp{57\% Positive\;} \posp{(69\% Positive\;)} & ... a hollow \textbf{joke (giggle)} told by a cinematic gymnast having too much fun embellishing the misanthropic tale to actually engage it .   \\
Vulnerable & \negp{56\% Negative} \posp{(56\% Positive\;)} & ... a hollow \textbf{joke (giggle)} told by a cinematic gymnast having too much fun embellishing the misanthropic tale \adv{cannot} actually engage it .   \\
\midrule
Original & \negp{73\% Negative} \negp{(56\% Negative)} & the \textbf{cold (colder)} turkey would 've been a far better title .   \\
Vulnerable & \negp{61\% Negative} \posp{(62\% Positive\;)} & the \textbf{cold (colder)} turkey \adv{might} 've been a far better title .   \\
\midrule
Original & \negp{70\% Negative} \negp{(65\% Negative)} & it 's just disappointingly superficial -- a movie that has all the elements necessary to be a fascinating , involving character study , but never does more than scratch the \textbf{shallow (surface)} .   \\
Vulnerable & \negp{52\% Negative} \posp{(55\% Positive\;)} & it 's just disappointingly \adv{short} -- a movie that has all the elements necessary to be a fascinating , involving character study , but never does more than scratch the \textbf{shallow (surface)} .   \\
\midrule
Original & \negp{79\% Negative} \negp{(72\% Negative)} & schaeffer has to find some hook on which to hang his persistently useless movies , and it might as well be the \textbf{resuscitation (revival)} of the middle-aged character .   \\
Vulnerable & \negp{57\% Negative} \posp{(57\% Positive\;)} & schaeffer has to find some hook on which to hang his persistently \adv{entertaining} movies , and it might as well be the \textbf{resuscitation (revival)} of the middle-aged character .   \\
\midrule
Original & \posp{64\% Positive\;} \posp{(58\% Positive\;)} & the primitive force of this film seems to bubble up from the vast collective memory of the \textbf{combatants (militants)} .   \\
Vulnerable & \posp{52\% Positive\;} \negp{(53\% Negative)} & the primitive force of this film seems to bubble \adv{down} from the vast collective memory of the \textbf{combatants (militants)} .   \\
\midrule
Original & \posp{64\% Positive\;} \posp{(74\% Positive\;)} & on this \textbf{troublesome (tricky)} topic , tadpole is very much a step in the right direction , with its blend of frankness , civility and compassion .   \\
Vulnerable & \negp{55\% Negative} \posp{(56\% Positive\;)} & on this \textbf{troublesome (tricky)} topic , tadpole is very much a step in the right direction , \adv{losing} its blend of frankness , civility and compassion .   \\
\midrule
Original & \posp{74\% Positive\;} \posp{(60\% Positive\;)} & if you 're \textbf{hard (laborious)} up for raunchy college humor , this is your ticket right here .   \\
Vulnerable & \posp{60\% Positive\;} \negp{(57\% Negative)} & if you 're \textbf{hard (laborious)} up for raunchy college humor , this is your ticket \adv{holder} here .   \\
\midrule
Original & \posp{94\% Positive\;} \posp{(97\% Positive\;)} & a fast , funny , highly \textbf{fun (enjoyable)} movie .   \\
Vulnerable & \negp{54\% Negative} \posp{(65\% Positive\;)} & a \adv{dirty} , \adv{violent} , highly \textbf{fun (enjoyable)} movie .   \\
\midrule
Original & \posp{86\% Positive\;} \posp{(88\% Positive\;)} & good old-fashioned slash-and-hack is \textbf{back (backwards)} !   \\
Vulnerable & \negp{52\% Negative} \posp{(55\% Positive\;)} & \adv{a} old-fashioned slash-and-hack is \textbf{back (backwards)} !   \\

\bottomrule
\end{tabularx}